\documentclass[twocolumn]{article}

\usepackage{arxiv}

\usepackage{multicol}
\usepackage[utf8]{inputenc} 
\usepackage[T1]{fontenc}    
\usepackage{hyperref}       
\usepackage{url}            
\usepackage{booktabs}       
\usepackage{amsfonts}       
\usepackage{nicefrac}       
\usepackage{microtype}      
\usepackage{graphicx}
\usepackage{doi}

\usepackage[
    backend=biber,
    bibstyle=ieee,
    citestyle=numeric-comp,
    hyperref=true,
    date=year,
    bibencoding=inputenc,
    uniquename=init,
    giveninits=true
]{biblatex}
\usepackage{longtable}
\usepackage{amsmath}
\usepackage{siunitx}
\usepackage{adjustbox}
\usepackage{subcaption}
\usepackage{acro}
\usepackage{xspace}
\usepackage{upgreek}
\usepackage{algorithm}
\usepackage{algorithmicx}
\usepackage{algpseudocode}
\usepackage{algcompatible}
\usepackage{relsize}
\usepackage{nth}
\usepackage[shortcuts]{extdash}
\usepackage{pifont}
\usepackage{mathtools}
\usepackage{newtxmath}

\newcommand{\cmark}{\ding{51}}%
\newcommand{\xmark}{\ding{55}}%
\newcommand{\autopq}{\texttt{AutoPQ}\xspace}
\newcommand{\autopqdef}{\texttt{AutoPQ\-/default}\xspace}
\newcommand{\autopqadv}{\texttt{AutoPQ\-/advanced}\xspace}

\makeatletter
\renewcommand{\ALG@beginalgorithmic}{\small}
\makeatother
\algrenewcommand\alglinenumber[1]{\footnotesize #1:}

\makeatletter
\def\ALG@special@indent{%
	\ifdim\ALG@thistlm=0pt\relax
	\hskip-\leftmargin
	\else
	\hskip\ALG@thistlm
	\fi
}
\newcommand{\INDENT}{\item[]\noindent\ALG@special@indent}
\newcommand{\INDENTAFTERFOR}{\item[]\noindent\ALG@special@indent\;\;\;\,}
\makeatother

\makeatletter
\newcommand\fs@norules{\def\@fs@cfont{\bfseries}\let\@fs@capt\floatc@ruled
	\def\@fs@pre{}%
	\def\@fs@post{}%
	\def\@fs@mid{\kern3pt}%
	\let\@fs@iftopcapt\iftrue}
\makeatother
\restylefloat{algorithm}

\renewcommand{\COMMENT}[1]{\textit{\scriptsize{\# #1}}}

\newcommand\ie{i.\,e.\xspace}
\newcommand\eg{e.\,g.\xspace}
\DeclareMathSymbol{\shortminus}{\mathbin}{AMSa}{"39}

\DeclareAcronym{PDF}{
    short = {PDF},
    long  = {Probability Density Function},
    tag = {acronyms}
}
\DeclareAcronym{PI}{
    short = {PI},
    long  = {Prediction Interval},
    tag = {acronyms}
}
\DeclareAcronym{HPO}{
    short = {HPO},
    long  = {Hyperparameter Optimization},
    tag = {acronyms}
}
\DeclareAcronym{CASH}{
    short = {CASH},
    long  = {Combined Algorithm Selection and Hyperparameter optimization},
    tag = {acronyms}
}
\DeclareAcronym{QRNN}{
    short = {QRNN},
    long  = {Quantile Regression Neural Network},
    tag = {acronyms}
}
\DeclareAcronym{PL}{
    short = {PL},
    long  = {Pinball Loss},
    tag = {acronyms}
}
\DeclareAcronym{NNQF}{
    short = {NNQF},
    long  = {Nearest Neighbor Quantile Filter},
    tag = {acronyms}
}
\DeclareAcronym{ML}{
    short = {ML},
    long  = {Machine Learning},
    tag = {acronyms}
}
\DeclareAcronym{MLP}{
    short = {MLP},
    long  = {MultiLayer Perceptron},
    tag = {acronyms}
}
\DeclareAcronym{SVR}{
    short = {SVR},
    long  = {Support Vector Regression},
    tag = {acronyms}
}
\DeclareAcronym{RF}{
    short = {RF},
    long  = {Random Forest},
    tag = {acronyms}
}
\DeclareAcronym{GBM}{
    short = {GBM},
    long  = {Gradient Boosting Machine},
    tag = {acronyms}
}
\DeclareAcronym{LASSO}{
    short = {LASSO},
    long  = {Least Absolute Shrinkage and Selection Operator},
    tag = {acronyms}
}
\DeclareAcronym{DL}{
    short = {DL},
    long  = {Deep Learning},
    tag = {acronyms}
}
\DeclareAcronym{SM}{
    short = {SM},
    long  = {Statistical Modeling},
    tag = {acronyms}
}
\DeclareAcronym{NHiTS}{
    short = {N-HiTS},
    long  = {Neural Hierarchical Interpolation for Time Series},
    tag = {acronyms}
}
\DeclareAcronym{DeepAR}{
    short = {DeepAR},
    long  = {Deep AutoRegression},
    tag = {acronyms}
}
\DeclareAcronym{CRPS}{
    short = {CRPS},
    long  = {Continuous Ranked Probability Score},
    tag = {acronyms}
}
\DeclareAcronym{MAQD}{
    short = {MAQD},
    long  = {Mean Absolute Quantile Deviation},
    tag = {acronyms}
}
\DeclareAcronym{MSE}{
    short = {MSE},
    long  = {Mean Squared Error},
    tag = {acronyms}
}
\DeclareAcronym{MRMR}{
    short = {MRMR},
    long  = {Minimum Redundancy Maximum Relevance},
    tag = {acronyms}
}
\DeclareAcronym{CDF}{
    short = {CDF},
    long  = {Cumulative Distribution Function},
    tag = {acronyms}
}
\DeclareAcronym{AR}{
    short = {AR},
    long  = {AutoRegression},
    tag = {acronyms}
}
\DeclareAcronym{RNN}{
    short = {RNN},
    long  = {Recurrent Neural Network},
    tag = {acronyms}
}
\DeclareAcronym{ANN}{
    short = {ANN},
    long  = {Artificial Neural Network},
    tag = {acronyms}
}
\DeclareAcronym{TBATS}{
    short = {TBATS},
    long  = {Trigonometric seasonality, Box-Cox transformation, ARMA errors, Trend, Seasonal components},
    tag = {acronyms}
}
\DeclareAcronym{ETS}{
    short = {ETS},
    long  = {Error Trend Seasonality},
    tag = {acronyms}
}
\DeclareAcronym{BO}{
    short = {BO},
    long  = {Bayesian Optimization},
    tag = {acronyms}
}
\DeclareAcronym{EA}{
    short = {EA},
    long  = {Evolutionary Algorithm},
    tag = {acronyms}
}
\DeclareAcronym{ARIMA}{
    short = {ARIMA},
    long  = {AutoRegressive Integrated Moving Average},
    tag = {acronyms}
}
\DeclareAcronym{TES}{
    short = {TES},
    long  = {Triple Exponential Smoothing},
    tag = {acronyms}
}
\DeclareAcronym{BATS}{
    short = {BATS},
    long  = {Box-Cox transformation, ARMA errors, Trend, Seasonal components},
    tag = {acronyms}
}
\DeclareAcronym{NBEATS}{
    short = {N-BEATS},
    long  = {Neural Basis Expansion Analysis for interpretable Time Series forecasting},
    tag = {acronyms}
}
\DeclareAcronym{TFT}{
    short = {TFT},
    long  = {Temporal Fusion Transformer},
    tag = {acronyms}
}
\DeclareAcronym{WP}{
    short = {WP},
    long  = {Wind Power},
    tag = {acronyms}
}
\DeclareAcronym{PV}{
    short = {PV},
    long  = {PhotoVoltaic},
    tag = {acronyms}
}
\DeclareAcronym{cINN}{
    short = {cINN},
    long  = {conditional Invertible Neural Network},
    tag = {acronyms}
}
\DeclareAcronym{HPC}{
    short = {HPC},
    long  = {High-Performance Computing},
    tag = {acronyms}
}
\DeclareAcronym{sARIMAX}{
    short = {sARIMAX},
    long  = {seasonal AutoRegressive Integrated Moving Average with eXternal input},
    tag = {acronyms}
}
\DeclareAcronym{XGB}{
    short = {XGB},
    long  = {eXtreme Gradient Boosting},
    tag = {acronyms}
}
\DeclareAcronym{TPE}{
    short = {TPE},
    long  = {Tree Parzen Estimator},
    tag = {acronyms}
}
\DeclareAcronym{PK}{
    short = {PK},
    long  = {Prior Knowledge},
    tag = {acronyms}
}
\DeclareAcronym{GPU}{
    short = {GPU},
    long  = {Graphics Processing Unit},
    tag = {acronyms}
}
\DeclareAcronym{CPU}{
    short = {CPU},
    long  = {Central Processing Unit},
    tag = {acronyms}
}
\DeclareAcronym{OPSD}{
    short = {OPSD},
    long  = {Open Power System Data},
    tag = {acronyms}
}
\DeclareAcronym{GCP}{
    short = {GCP},
    long  = {Grid Connection Point},
    tag = {acronyms}
}
\DeclareAcronym{GEFCom}{
    short = {GEFCom},
    long  = {Global Energy Forecasting Competition},
    tag = {acronyms}
}
\DeclareAcronym{ECMWF}{
    short = {ECMWF},
    long  = {European Centre for Medium-Range Weather Forecasts},
    tag = {acronyms}
}
\DeclareAcronym{NWP}{
    short = {NWP},
    long  = {Numerical Weather Prediction},
    tag = {acronyms}
}
\DeclareAcronym{GHI}{
    short = {GHI},
    long  = {Global Horizontal Irradiance},
    tag = {acronyms}
}
\DeclareAcronym{RAM}{
    short = {RAM},
    long  = {Random-Access Memory},
    tag = {acronyms}
}
\DeclareAcronym{GAN}{
    short = {GAN},
    long  = {Generative Adversarial Network},
    tag = {acronyms}
}

\title{\autopq: Automating Quantile estimation from Point forecasts in the context of sustainability}
\date{} 					

\author{
    \href{https://orcid.org/0000-0002-9320-5341}{\includegraphics[scale=0.06]{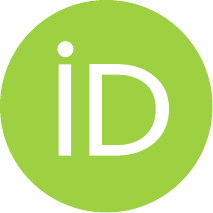}\hspace{1mm}
    Stefan Meisenbacher}\\
	Institute for Automation and Applied Informatics\\
	Karlsruhe Institute of Technology\\
	Eggenstein-Leopoldshafen, 76344, Germany\\
	\texttt{stefan.meisenbacher@kit.edu}\\
	\And
    \href{https://orcid.org/0000-0002-9197-1739}{\includegraphics[scale=0.06]{orcid.pdf}\hspace{1mm}
    Kaleb Phipps}\\
	Scientific Computing Center (SCC)\\
	Karlsruhe Institute of Technology\\
	Eggenstein-Leopoldshafen, 76344, Germany\\
	\And
    \href{https://orcid.org/0000-0002-3707-499X}{\includegraphics[scale=0.06]{orcid.pdf}\hspace{1mm}
    Oskar Taubert}\\
	Scientific Computing Center (SCC)\\
	Karlsruhe Institute of Technology\\
	Eggenstein-Leopoldshafen, 76344, Germany\\
	\And
    \href{https://orcid.org/0000-0001-9648-4385}{\includegraphics[scale=0.06]{orcid.pdf}\hspace{1mm}
    Marie Weiel}\\
	Scientific Computing Center (SCC)\\
	Karlsruhe Institute of Technology\\
	Eggenstein-Leopoldshafen, 76344, Germany\\
	\And
    \href{https://orcid.org/0000-0002-2233-1041}{\includegraphics[scale=0.06]{orcid.pdf}\hspace{1mm}
    Markus Götz}\\
	Scientific Computing Center (SCC)\\
	Karlsruhe Institute of Technology\\
	Eggenstein-Leopoldshafen, 76344, Germany\\
	\And
    \href{https://orcid.org/0000-0001-9100-5496}{\includegraphics[scale=0.06]{orcid.pdf}\hspace{1mm}
    Ralf Mikut}\\
	Institute for Automation and Applied Informatics\\
	Karlsruhe Institute of Technology\\
	Eggenstein-Leopoldshafen, 76344, Germany\\
	\And
    \href{https://orcid.org/0000-0002-3572-9083}{\includegraphics[scale=0.06]{orcid.pdf}\hspace{1mm}
    Veit Hagenmeyer}\\
	Institute for Automation and Applied Informatics\\
	Karlsruhe Institute of Technology\\
	Eggenstein-Leopoldshafen, 76344, Germany\\
}

\renewcommand{\headeright}{Meisenbacher \textit{et al.}}
\renewcommand{\undertitle}{}
\renewcommand{\shorttitle}{\autopq}

\hypersetup{
pdftitle={AutoPQ: Automated point forecast-based quantile forecasts},
pdfsubject={automated point forecast-based quantile forecasts},
pdfauthor={Meisenbacher et al.},
pdfkeywords={
    Probabilistic time series forecasting,
    Uncertainty quantification,
    AutoML,
    Energy Consumption
    },
}

\begin{document}

\twocolumn[
    \begin{@twocolumnfalse}
        \maketitle
        \begin{abstract}
            Optimizing smart grid operations relies on critical decision-making informed by uncertainty quantification, making probabilistic forecasting a vital tool.
            Designing such forecasting models involves three key challenges:
            accurate and unbiased uncertainty quantification,
            workload reduction for data scientists during the design process,
            and limitation of the environmental impact of model training.
            In order to address these challenges, we introduce \autopq, a novel method designed to automate and optimize probabilistic forecasting for smart grid applications.
            \autopq enhances forecast uncertainty quantification by generating quantile forecasts from an existing point forecast by using a \ac{cINN}.
            \autopq also automates the selection of the underlying point forecasting method and the optimization of hyperparameters, ensuring that the best model and configuration is chosen for each application.
            For flexible adaptation to various performance needs and available computing power, \autopq comes with a default and an advanced configuration, making it suitable for a wide range of smart grid applications.
            Additionally, \autopq provides transparency regarding the electricity consumption required for performance improvements.
            We show that \autopq outperforms state-of-the-art probabilistic forecasting methods while effectively limiting computational effort and hence environmental impact.
            Additionally and in the context of sustainability, we quantify the electricity consumption required for performance improvements.
        \end{abstract}
    \end{@twocolumnfalse}
]

\clearpage
\section{Introduction}
Uncertainty quantification is necessary for making informed decisions in smart grid applications, rendering probabilistic time series forecasts essential~\cite{Khajeh2022}.
Exemplary downstream applications are stochastic power flow optimizations~\cite{Summers2015,Mieth2018}, smart charging of stationary battery systems and electric vehicles~\cite{Appino2018,Henni2022,Phipps2023b,Huber2020}, and economic dispatch~\cite{Zhang2021,Han2021}.
As the number of applications increases, automating the design process of probabilistic time series forecasting models is necessary to keep pace with the growing demand.

This process involves three major challenges: First, a forecast's inherent uncertainty must be quantified in an unbiased and accurate manner.
However, many recently published methods~\cite{Oreshkin2020, Lim2021b, Challu2023} only provide point forecasts~\cite{Petropoulos2022}.
Methods to generate a probabilistic forecast from a point forecast exist but have limitations, \eg they assume a \ac{PDF}~\cite{Hyndman2021, Camporeale2019}), only estimate \acp{PI}~\cite{Williams1971, Stankeviciute2021}, or rely on the residuals of the point forecast~\cite{Wang2020b} instead of representing the uncertainty of the process itself.
Although direct probabilistic methods can partially overcome these limitations, they fail to leverage the many existing and well-established forecasting models already in use~\cite{Phipps2024}.
\\
Second, given that the forecast quality is sensitive to different model design decisions~\cite{Hutter2019}, no forecasting method can excel in all tasks (no\-/free\-/lunch theorem), and the quality requirements depend on the downstream application.
That is why automatically selecting the best\-/performing method is absolutely crucial~\cite{Phipps2024}.
In addition, performance\-/critical smart grid applications often demand further model enhancement by \ac{HPO}.
However, existing \ac{CASH} methods in time series forecasting are limited to point forecasting~\cite{Raetz2019, Shah2021, Deng2023} or are subject to the constraints outlined in the first challenge~\cite{Shchur2023}.
\\
Third, methodological advances in probabilistic forecasting must quantify the environmental impact of employing such an advanced model in a standardized, quantifiable, and comparable manner~\cite{Debus2023}.
Specifically, smart grid applications utilizing forecasts must ensure that the electricity consumption required for forecasting model design remains at a reasonable level.
This is crucial because excessive electricity consumption, particularly for computationally intensive modern \ac{DL} methods, can undermine sustainability efforts by increasing the overall carbon footprint.
Despite this need, most existing probabilistic forecasting methods only specify the computing hardware employed~\cite{Salinas2020,Rangapuram2021,Shchur2023} but fail to report the associated electricity consumption.

To tackle these challenges, we introduce \autopq, a novel method for automated and electricity consumption-aware quantile forecasts from point forecasts.
Our key contributions include:
\begin{itemize}
    \item \autopq generates quantile forecasts without requiring prior information about the underlying distribution by i) using established point forecasting methods (unknown distribution), ii) mapping the point forecast into the latent space of a \acf{cINN}~\cite{Heidrich2023} (known and tractable distribution), and iii) sampling in the neighborhood of the point forecast's latent space representation~\cite{Phipps2024}.
    \item \autopq enhances~\cite{Phipps2024} and automates model design decisions by \textit{a}) optimizing the variance for sampling in the latent space, \textit{b}) selecting the best\-/performing point forecasting method, and \textit{c}) optimizing its hyperparameters.
    \item In order to accommodate various computing systems and performance requirements, we introduce \autopqdef (automating \textit{a} and \textit{b}) for general\-/purpose computing systems to provide high\-/quality probabilistic forecasts, and \autopqadv (automating \textit{a}, \textit{b}, and \textit{c}) for \ac{HPC} systems to increase the forecasting quality further, as required by smart grid applications with high decision costs.
    Additionally, the electricity consumption of \autopqdef and \autopqadv is reported and compared in relation to their achievable performance.
\end{itemize}

The remainder of the paper is structured as follows:
\autoref{sec:related_work} provides an overview of relevant related work.
In \autoref{sec:autopq} we then present \autopq, which we further evaluate in \autoref{sec:evaluation}.
The results of our evaluation are discussed in \autoref{sec:discussion}.
Finally, \autoref{sec:conclusion} gives a conclusion and an outlook.

\section{Related work}
\label{sec:related_work}
This section analyzes related work with a particular focus on the three challenges stated above, \ie, uncertainty quantification in time series forecasting, design process automation, and quantification of energy consumption required for performance improvements.
\autoref{tab:autopq_related-work} gives an overview of the analyzed literature.

\begin{table*}
    \caption{
        Overview of existing time series forecasting methods in terms of uncertainty quantification, automated design, and reporting the electricity consumption for the model design.
    }
    \label{tab:autopq_related-work}
    \centering
    \begin{adjustbox}{max width=\textwidth}
\begin{tabular}{llccccc}
\toprule
\textbf{Reference} & \textbf{Category} & \multicolumn{3}{c}{\textbf{Uncertainty quantification}} & \textbf{Automated} & \textbf{Electricity} \\
      &       & \textbf{Quantiles/\acsp{PI}} & \textbf{\acs{PDF}} & \textbf{Scenarios} & \textbf{design} & \textbf{consumption} \\
\midrule[\heavyrulewidth]
\cite{Cannon2011, GonzalezOrdiano2020} & Direct probabilistic forecast & \cmark     & \xmark     & \xmark     & \xmark     & \xmark \\
\cite{Salinas2020} & Direct probabilistic forecast & \cmark     & \cmark     & \xmark     & \xmark     & \xmark \\
\cite{Rangapuram2021} & Direct probabilistic forecast & \cmark     & \cmark     & \xmark     & \acs{HPO}  & \xmark \\
\cite{Wen2019,Rasul2021,Jamgochian2022,Arpogaus2023,Fanfarillo2021,Zhang2020,Ge2020,Dumas2022,Cramer2023} & Direct probabilistic forecast & \cmark     & \cmark     & \cmark     & \xmark     & \xmark \\
\cite{Hyndman2021,Williams1971,Stankeviciute2021,Chernozhukov2021,Zaffran2022} & Point forecast-based probabilistic forecast & \cmark     & \xmark     & \xmark     & \xmark     & \xmark \\
\cite{Wang2020b} & Point forecast-based probabilistic forecast & \cmark     & \cmark     & \cmark     & \xmark     & \xmark \\
\cite{Camporeale2019} & Point forecast-based probabilistic forecast & \cmark     & \cmark     & \xmark     & \xmark     & \xmark \\
\cite{Hyndman2002,Hyndman2008,DeLivera2011,Maldonado2019,Valente2020,Fan2019,Barros2021,Wu2021,Kong2017,AlMamun2019} & Point forecast & \xmark     & \xmark     & \xmark     & \acs{HPO}  & \xmark \\
\cite{Raetz2019,Shah2021,Deng2023} & Point forecast & \xmark     & \xmark     & \xmark     & \acs{CASH} & \xmark \\
\cite{Shchur2023} & Point forecast-based and direct probabilistic forecast & \cmark     & \xmark     & \xmark     & \acs{CASH} & \xmark \\
\bottomrule
AutoPQ-default & Point forecast-based probabilistic forecast & \cmark     & \cmark     & \cmark     & \acs{HPO}  & \cmark \\
AutoPQ-advanced & Point forecast-based probabilistic forecast & \cmark     & \cmark     & \cmark     & \acs{CASH} & \cmark \\
\end{tabular}%

    \end{adjustbox}
    \scriptsize\\[1mm]
    \acf{HPO}, \acf{CASH}
\end{table*}

\subsection{Probabilistic time series forecasting}
Probabilistic time series forecasting methods can be categorized into direct probabilistic methods and probabilistic methods based on point forecasts.

\paragraph{Direct probabilistic forecasting methods.}
Direct probabilistic forecasting methods aim to learn the uncertainty of the forecast during model training.
Depending on how a forecast's uncertainty is represented, they can be divided into quantile\-/based, distribution\-/based, and normalizing flow\-/based forecasts.

Quantile\-/based forecasting methods are trained to represent certain quantiles.
These quantiles can then be applied to derive \acp{PI} with a specific conditional probability that the target value lies within the interval.
For instance, a \ac{QRNN}~\cite{Cannon2011} uses the \ac{PL} function during training such that the resulting forecast represents a specific quantile.
Multiple \acp{QRNN} representing different quantiles can be trained to forecast this set of quantiles.
A different approach is the \ac{NNQF}~\cite{GonzalezOrdiano2020} that modifies the training dataset based on similarity in the target variable to determine a set of quantiles.
Afterward, the method trains a \ac{ML}\-/based method on the modified data to forecast this set of quantiles.

However, both \acp{QRNN} and the \ac{NNQF} can only forecast quantiles, from which \acp{PI} can be derived, but do not provide the probability distribution.
\\
The full \ac{PDF} or \ac{CDF} can be obtained from distribution\-/based forecasts.
A well\-/known example is \ac{DeepAR}~\cite{Salinas2020}, an \ac{AR}\-/based \ac{RNN} trained to learn the parameters of a given \ac{PDF} and generate probabilistic forecasts by sampling from it.
In~\cite{Rangapuram2021}, the authors extend \ac{DeepAR} to make probabilistic hierarchical forecasts.
However, both methods require assuming the underlying \ac{PDF}, which is limiting if the data does not follow a standard parametric distribution.
\\
This disadvantage is overcome by normalizing flow\-/based probabilistic forecasts, which learn a bijective mapping from the unknown \ac{PDF} of the data to a known and tractable \ac{PDF}.
Typically, normalizing flows are used to enhance existing probabilistic forecasting methods, such as \acp{QRNN}~\cite{Wen2019}, \acp{RNN}~\cite{Rasul2021}, Gaussian mixture models~\cite{Jamgochian2022}, and Bernstein polynomials~\cite{Arpogaus2023}.
They can also be applied to directly generate probabilistic forecasts for various domains, including atmospheric variables~\cite{Fanfarillo2021}, renewable energy generation~\cite{Dumas2022}, electrical load~\cite{Zhang2020, Ge2020,Dumas2022}, and electricity prices~\cite{Cramer2023}.
Although these methods are effective, many do not consider exogenous features like weather variables to condition the \ac{PDF}, \ie, they assume that the learned distribution remains constant.

All direct probabilistic forecasting methods are limited by their inability to generate probabilistic forecasts from available, well\-/designed point forecasts~\cite{Petropoulos2022}.

\paragraph{Point forecast-based probabilistic forecasting methods.}
Point forecast-based probabilistic forecasting methods quantify forecast uncertainty using a separate approach for uncertainty estimation~\cite{Smith2013}.
In this context, most uncertainty quantification methods rely on the residuals between point forecasts and observed values in a validation dataset to estimate \acp{PI}.
The assumptions regarding the distribution of the residuals differ between methods.
For instance, they can be assumed to follow a Gaussian distribution~\cite{Hyndman2021}, an empirical distribution~\cite{Williams1971}, or an empirical non\-/conformity score distribution~\cite{Stankeviciute2021, Chernozhukov2021, Zaffran2022}.
The latter calculates a critical non\-/conformity score for the desired significance level of the \ac{PI} by assessing how unusual a residual is compared to other residuals in the validation dataset.
However, these methods can only estimate \acp{PI} and do not provide the full \ac{PDF}.

Beyond the above approaches, there are enhancements in residual\-/based uncertainty quantification that leverage \ac{ML} techniques.
In~\cite{Wang2020b}, the authors train a \ac{GAN} on the residuals of a point forecast, which is subsequently used to generate residual scenarios and quantify the point forecast's uncertainty.
In~\cite{Camporeale2019}, a Gaussian distribution of residuals is assumed, and an \ac{ANN} is trained to forecast the standard deviation of these residuals.
The estimated standard deviation is then used to quantify the point forecast's uncertainty.
As the uncertainty estimation models are trained on the point forecast's residuals, both approaches directly depend on the point forecast.
Consequently, when comparing multiple candidate point forecasting methods and hyperparameter configurations, a separate uncertainty estimation model must be trained for each of them.

\subsection{Automated time series forecasting}
Automated time series forecasting methods can be classified into two main categories: \acf{HPO} and \acf{CASH}.
While automated point forecasting utilizing \ac{HPO}~\cite{Hyndman2002,Hyndman2008,DeLivera2011,Maldonado2019,Valente2020,Fan2019,Barros2021,Wu2021,Kong2017,AlMamun2019} and \ac{CASH}~\cite{Raetz2019,Shah2021,Deng2023} has been extensively researched, there is a significant gap in studies that focus on automated probabilistic forecasting~\cite{Meisenbacher2022}.

\paragraph{Hyperparameter optimization.}
The purpose of \ac{HPO} is to find a forecasting method's configuration hat delivers the best\-/possible performance for a given forecasting task.

Simple grid search\-/based \ac{HPO} has been proposed for point forecasting methods within the \ac{SM} family, such as \ac{ETS}~\cite{Hyndman2002}, \ac{sARIMAX}~\cite{Hyndman2008}, and \ac{TBATS}~\cite{DeLivera2011}.
Grid search is also utilized within the \ac{ML} family, \eg, for \ac{HPO} of point forecasting methods based on \ac{SVR}~\cite{Maldonado2019, Valente2020}.
While effective for small configuration spaces, higher-dimensional spaces require \ac{HPO} algorithms that explore the configuration space more efficiently.
For example, random search is more efficient than grid search~\cite{Bergstra2012} and is widely applied for \ac{HPO} of \ac{ML}\-/based point forecasting methods such as the \ac{GBM}~\cite{Barros2021} and the \ac{MLP}~\cite{Fan2019}.
For a directed search of complex configuration spaces, such as for methods of the \ac{DL} family, gradient\-/based search methods~\cite{Wu2021}, \ac{BO}~\cite{Kong2017}, and \acp{EA}~\cite{AlMamun2019} are used.
Aside from point forecasting methods, these \ac{HPO} techniques can also be applied to \ac{DL}\-/based direct probabilistic forecasting methods, as demonstrated in~\cite{Rangapuram2021}.
Although \ac{HPO} of a single forecasting method is effective, it neglects the no\-/free lunch theorem, stating that no single forecasting method excels at all forecasting tasks.

\paragraph{Combined algorithm selection and hyperparameter optimization.}
To select the best\-/performing forecasting method with optimized hyperparameters, several \ac{CASH} methods are available for point forecasting.
In~\cite{Raetz2019}, the authors take different \ac{ML}\-/based forecasting methods into account, including \ac{LASSO}, \ac{RF}, \ac{XGB}, \ac{MLP}, and \ac{SVR}, and use \ac{BO} for solving the \ac{CASH} problem.
Similar approaches exist in~\cite{Shah2021} and~\cite{Deng2023}, additionally considering point forecasting methods of the \ac{SM} family (\eg, \ac{ARIMA}, \ac{TES}, and \ac{BATS}) and the \ac{DL} family (\eg, \ac{NHiTS} and \ac{TFT}).
Apart from the application to point forecasting, \ac{CASH} methods are also used for direct and point forecast\-/based probabilistic forecasting~\cite{Shchur2023}.
In their \ac{CASH} problem, the authors consider several direct probabilistic forecasting methods, like \ac{DeepAR}, as well as point forecasting methods of the \ac{SM} and \ac{ML} family using conformal \acp{PI} to derive probabilistic forecasts.
Despite their contribution to addressing the under\-/researched area of \ac{CASH} for probabilistic forecasting, the applied uncertainty quantification methods are still subject to the aforementioned limitations, \ie, require assuming the underlying \ac{PDF} and do not take into account the conditioning of the \ac{PDF} by exogenous features.

\subsection{Computing effort for methodological advances}
Regarding the quantification of efforts required for performance improvements, the above\-/reviewed related works exhibit significant shortcomings.
While some of the studies specify the used hardware and required training duration~\cite{Shchur2023,Raetz2019,Deng2023,Maldonado2019,Wu2021,Fanfarillo2021,Dumas2022,Salinas2020,GonzalezOrdiano2020}, estimating the corresponding energy consumption from this information involves considerable effort, with inherent uncertainty.
Yet, the electricity consumption is an essential metric as it is independent of data center\-/specific or geopolitical factors at the time of measurement~\cite{Debus2023}.
Once electricity consumption is quantified, other metrics, such as the carbon footprint or the monetary costs, can be derived using assumed or local conditions.

\section{\autopq}
\label{sec:autopq}
\begin{figure*}[!h]
    \centering
    \includegraphics[width=\linewidth]{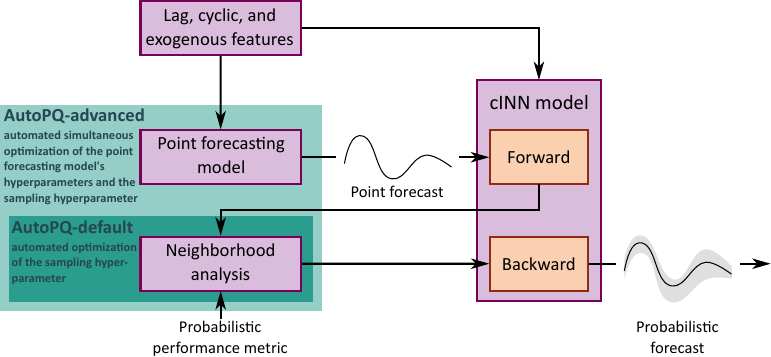}
    \caption{
        Overview of \autopq:
        Lag features, seasonal features, and exogenous features are selected and used as inputs by a point forecasting model to generate a point forecast in an unknown distribution.
        This point forecast and the features are combined in a \acs{cINN}, resulting in a representation of the forecast in a known and tractable distribution.
        The neighborhood of this representation is analyzed to determine how to include uncertainty information.
        Finally, with the backward pass through the \acs{cINN}, the uncertainty is mapped back to the unknown distribution to generate the probabilistic forecast.
        Automation methods are highlighted in green.
        }
    \label{fig:autopq_method}
\end{figure*}

To address the three key challenges in the automated design of probabilistic time series forecasting models, we present \autopq, an innovative method for electricity consumption-aware quantile forecasting based on point forecasts.
The idea behind \autopq is to use a \ac{cINN}~\cite{Heidrich2023} to generate a probabilistic forecast from any arbitrary point forecast~\cite{Phipps2024} while automatically making corresponding design decisions to enhance probabilistic performance.
In the following section, we describe i) the generation of such a probabilistic forecast, ii) the corresponding design decisions, and iii) the automation of these decisions, highlighted green in \autoref{fig:autopq_method}.
To accommodate various computing systems and performance requirements, we propose two variants of \autopq: \autopqdef, designed for general\-/purpose computing systems and capable of providing high\-/quality probabilistic forecasts~\cite{Phipps2024}, and \autopqadv, which requires \ac{HPC} systems\footnote{
    In this paper, we refer to servers for parallel computing with multiple nodes and \ac{GPU} acceleration as \ac{HPC} systems.
} to further increase forecasting quality for downstream applications with high decision costs.

\subsection{Creation of the probabilistic forecast}
Generating a probabilistic forecast with a \ac{cINN} using a point forecast involves three steps~\cite{Phipps2024}.\footnote{
    For a detailed description, we refer to~\cite{Phipps2024}.
}

\paragraph{Creation of a point forecast in an unknown distribution.}
A point forecast is generated using an arbitrary point forecasting model $f_\text{p}\left(\cdot\right)$, which estimates future values $\hat{\mathbf{y}}$ of the target time series $\mathbf{y}$ using current and past values.
More precisely, the model $f_\text{p}\left(\cdot\right)$ makes a forecast $\hat{\mathbf{y}}\left[k \!+\! 1, \ldots, k \!+\! H\right]$ for a specified forecast horizon $H \in \mathbb{N}_1$ at origin $k$, taking into account historical values of $\mathbf{y}\left[k \!\shortminus\! H_1, \ldots, k\right]$, values of exogenous forecasts $\hat{\mathbf{X}}\left[k \!+\! 1,\ldots,k \!+\! H\right]$, and exogenous time series $\mathbf{X}\left[k \!\shortminus\! H_1, \ldots, k\right]$.
Since the target time series $\mathbf{y}$ contains realizations of $\hat{\mathbf{y}}$, a point forecast can be interpreted as a sample of the random variable $\mathrm{Y} \sim f_\mathrm{Y}\left(\mathbf{y}\right)$ in the realization space $\mathbb{Y}$ with an unknown $H$\-/dimensional \ac{PDF} $f_\mathrm{Y}\left(\mathbf{y}\right)$.

\paragraph{Representation of the point forecast in a known and tractable distribution.}
The point forecast $\hat{\mathbf{y}}\left[k \!+\! 1, \ldots, k \!+\! H\right]$ as a sample from the unknown distribution is represented in a space with a known and tractable $H$\-/dimensional \ac{PDF} $f_\mathrm{Z}\left(\mathbf{y}\right)$ using a \ac{cINN}.
The \ac{cINN} learns a conditional\footnote{
    The \ac{cINN} can include additional exogenous information into the mapping using conditional\-/affine coupling blocks~\cite{Ardizzone2019}. 
} and bijective function $g: \mathbb{Y} \rightarrow \mathbb{Z}$~\cite{Ardizzone2019} from the realization space $\mathbb{Y}$ with unknown distribution into the latent space $\mathbb{Z}$ with a multi\-/dimensional Gaussian distribution.
This bijective function can be used to quantify the point forecast's uncertainty by mapping it into the latent space and analyzing its neighborhood as described in the following.
Notably, the training of the \ac{cINN} is independent of the point forecasting model, which allows using one \ac{cINN} to generate probabilistic forecasts from multiple arbitrary point forecasts.

\paragraph{Neighborhood analysis of the point forecast's latent space representation.}
To obtain the point forecast's latent space representation, the point forecast $\hat{\mathbf{y}}\left[k \!+\! 1, \ldots, k \!+\! H\right]$ itself and additional exogenous information $\{\mathbf{y}\left[k \!\shortminus\! H_1, \ldots, k\right], \mathbf{X}\left[k \!\shortminus\! H_1, \ldots, k\right],\hat{\mathbf{X}}\left[k \!+\! 1,\ldots,k \!+\! H\right]\}$ are passed forward through the trained \ac{cINN}.
To quantify the point forecast's uncertainty, the neighborhood of this representation is analyzed.
Samples are generated in the Gaussian-distributed latent space with a sampling variance $\sigma$, resulting in a set of realizations that are similar but not identical to the original point forecast.
These samples represent the uncertainty in the neighborhood of the point forecast's latent space representation and can be passed backward through the \ac{cINN} to calculate the quantiles of the probabilistic forecast $\hat{\mathbf{y}}^{*}\left[k \!+\! 1, \ldots, k \!+\! H\right]$.\footnote{
    This is valid due to the equivalence of uncertainty in both spaces~\cite{Phipps2024}.
}

\subsection{Task-dependent design decisions}

Generating a quantile forecast requires specific design decisions tailored to the forecasting task.
These include feature engineering, selecting the point forecasting method and its hyperparameters $\boldsymbol{\uplambda}_\text{p}$, and choosing the sampling variance $\uplambda_\text{q} = \sigma$ for generating a quantile forecast with the \ac{cINN} based on the point forecast's latent space representation.

\paragraph{Lag, cyclic, and exogenous features.}
Additional explanatory variables can enrich the input space of both the point forecasting model and the \ac{cINN}.
Lag features provide past values for the model as historical context.
In the evaluation, point forecasting methods based on \ac{ML} regression estimators require the explicit definition of lag features
\begin{footnotesize}
\begin{equation}
    x_{\text{lag},H_1}\left[k\right] = x\left[k\!\shortminus\!H_1\right],
    \label{eq:concept_lag}
\end{equation}
\end{footnotesize}

unlike \ac{SM}- and \ac{DL}\-/based forecasting methods that consider past values implicitly.
Seasonal features represent cyclic relationships, such as the hour of the day or the month,
\begin{scriptsize}
\begin{align}
    {x_{\text{s12}}}\left[k\right]     = \sin \left(\frac{2 \pi \cdot \text{month}\left[k\right]}{12}\right),
    {x_{\text{c12}}}\left[k\right]     = \cos \left(\frac{2 \pi \cdot \text{month}\left[k\right]}{12}\right)
    \label{eq:concept_trig-12},\\
    {x_{\text{s24}}}\left[k\right]   = \sin \left(\frac{2 \pi \cdot \text{hour}\left[k\right]}{24}\right),
    \;\;\,
    {x_{\text{c24}}}\left[k\right]   = \cos \left(\frac{2 \pi \cdot \text{hour}\left[k\right]}{24}\right)
    ,\;\;
    \label{eq:concept_trig-24}
\end{align}
\end{scriptsize}

as well as work and non\-/work days:
\begin{footnotesize}
\begin{equation}
    x_\text{wd}\left[k\right] = \begin{cases*}
        1,  & if $\texttt{work day}(k)$ is True,\\
        0   & otherwise.\\
    \end{cases*}
    \label{eq:concept_workday-weekend-holiday}
\end{equation}
\end{footnotesize}

Since energy time series for smart grid applications contain predominantly daily, weekly, and yearly patterns~\cite{Heidrich2020, Giacomazzi2023, Richter2023}, we consider seasonal features in the evaluation.
Exogenous features are important if the target variable is influenced by external factors, such as weather conditions including temperature, humidity, wind speed, and solar irradiance.
We thus select exogenous features depending on existing additional variables of the datasets in our evaluation~\cite{Phipps2024}.

\paragraph{Point forecasting methods and hyperparameters.}
As the point forecasting method in the above-described \ac{cINN} approach significantly affects the probabilistic forecasting performance~\cite{Phipps2024}, we consider established methods from different forecasting method families in order to obtain a representative selection of methods that have proven effective and differ in their approach or architecture.
\begin{description}
    \item[\acf{SM}:] \acs{sARIMAX}, \acs{ETS}, \acs{TBATS},
    \item[\acf{ML}:] \acs{MLP}, \acs{SVR}, \acs{XGB},
    \item[\acf{DL}:] \acs{DeepAR}, \acs{NHiTS}, \acs{TFT}.
\end{description}
Since \ac{HPO} can enhance performance beyond a method's default configuration, we apply \ac{HPO} on the configuration spaces detailed in \autoref{tab:autopq_sm-configurations}, \ref{tab:autopq_ml-configurations}, and \ref{tab:autopq_dl-configurations} in the Appendix.

\paragraph{Sampling hyperparameter of the latent space.}
The hyperparameter $\uplambda_\text{q} = \sigma$ for generating samples from the point forecast's latent space representation controls the magnitude of quantified uncertainty in the forecast: 
A larger $\sigma$ includes more uncertainty and improves coverage rates, while a smaller $\sigma$ produces sharper probabilistic forecasts.
Both coverage rate and sharpness can be evaluated using probabilistic performance metrics, enabling \ac{HPO} to tailor the sampling hyperparameter $\uplambda_\text{q} = \sigma$ to the forecasting task ~\cite{Phipps2023a}. 
The considered value range is given in \autoref{tab:autopq_cinn-configurations} in the Appendix.

\subsection{Optimal design of the probabilistic forecast}
\autopq automates the design decisions outlined above by optimally configuring the forecasting model.
We propose \autopqdef for general\-/purpose computing systems to provide high\-/quality probabilistic forecasts, and \autopqadv for \ac{HPC} systems to enhance the performance further.

\subsection{\autopqdef for general\-/purpose computing systems}
\begin{figure*}[tb!]
    \centering
    \includegraphics{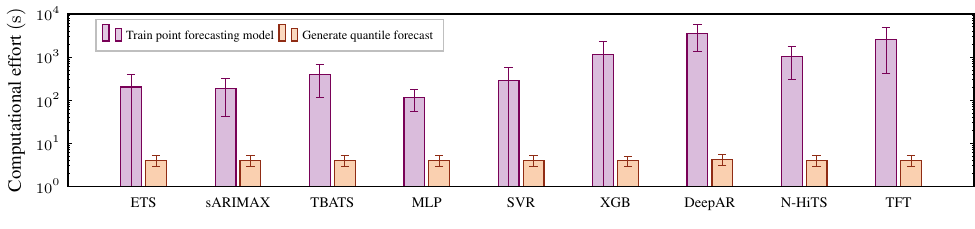}
    \caption{
        Comparison of the average computational effort in terms of runtime across the six datasets used in the evaluation: effort for training a point forecasting model with configuration $\uplambda_\text{p}$ and effort for generating a quantile forecast based on the point forecast using the \acs{cINN} with $\boldsymbol{\uplambda}_\text{q}$.
        Due to the significantly smaller computational effort of generating the quantile forecast, it is worthwhile to evaluate several $\uplambda_\text{q}$ for one $\boldsymbol{\uplambda}_\text{p}$ to properly balance the efforts.
        Note that for \acs{ETS}, \acs{SVR}, and \acs{XGB} the standard deviation is higher than the mean value.
    }
    \label{fig:autopq_computational-effort-total_bar-plot}
\end{figure*}

Given the high computational effort of \ac{HPO} for point forecasting methods illustrated in \autoref{fig:autopq_computational-effort-total_bar-plot}, \autopqdef only focuses on optimizing the sampling hyperparameter to provide high\-/quality point forecast\-/based quantile forecasts using general\-/purpose computing systems for the model design.


\paragraph{Hyperparameter optimization.}
Instead of manually exploring a given configuration space $\mathbf{\Lambda}_\text{q}$ to find the (estimated) optimal hyperparameter configuration $\uplambda_\text{q}^{\hat{\star}}$, we define the \ac{HPO} problem
\begin{equation}
    \uplambda_\text{q}^{\hat{\star}} = \min_{\boldsymbol{\uplambda} \in \mathbf{\Lambda}_\text{q}} Q\left(\hat{\mathbf{y}}(\uplambda_\text{q}), \mathbf{y}\right),
    \label{eq:autopq_objective_default}
\end{equation}
with an arbitrary probabilistic performance metric as objective function $Q$ and the hyperparameter $\uplambda_\text{q}$ for generating samples from the point forecast's latent space representation.

\paragraph{Optimization algorithm.}
Given that the objective function $Q$ \eqref{eq:autopq_objective_default} is potentially noisy and non\-/differentiable and we do not have a closed\-/form expression of the configuration space $\mathbf{\Lambda}_\text{q}$, we use the black\-/box optimizer \texttt{Hyperopt}~\cite{Bergstra2013}.
\texttt{Hyperopt} acts as the \ac{HPO} trial generator suggesting new candidate configurations to be assessed.
The suggestion is based on \ac{BO} using the \ac{TPE} as surrogate model, which balances exploration and exploitation of promising regions in the configuration space.

\paragraph{Early stopping.}
The optimization can be stopped once the objective function's value $Q$ has converged or reached a satisfactory value.
To terminate the \ac{BO} when additional trials are unlikely to improve $Q$,
we use early stopping upon $Q$ reaching a plateau, defined as the standard deviation of the five best trials falling below \num{0.0005}, with a patience of five trials.\footnote{
    Note that the assessment is performed on normalized data, \ie, the values of $Q$ across datasets are in similar ranges.
}

\paragraph{Forecasting method selection.}
Probabilistic forecasting performance depends on not only the hyperparameter $\uplambda_\text{q}$ but also the underlying point forecasting method.
In order to address this, \autopqdef selects the best\-/performing point forecasting method while optimizing the corresponding $\uplambda_\text{q}$.
First, all candidate point forecasting models are trained with default configurations $\boldsymbol{\uplambda}_\text{p,default}$.
A quantile forecast is then generated for each model, with the associated sampling hyperparameter $\uplambda_\text{q}$ optimized with the previously described \ac{HPO}.
Finally, the combination of point forecasting model and optimized sampling hyperparameter achieving the best score on the validation dataset is selected.

\subsubsection{\autopqadv for HPC systems}
\label{sssec:autopq_advanced}
In smart grid applications with high decision costs, improving probabilistic forecasting performance is vital.
\autopqadv addresses this requirement using \ac{HPC} systems for the model design as detailed in the following.

\paragraph{Hyperparameter optimization.}
Similarly to \autopqdef \eqref{eq:autopq_objective_default}, the \ac{HPO} problem
\begin{equation}
    \boldsymbol{\uplambda}^{\hat{\star}} = \min_{\boldsymbol{\uplambda} \in \mathbf{\Lambda}} Q\left(\hat{\mathbf{y}}(\boldsymbol{\uplambda}_\text{p}, \uplambda_\text{q}), \mathbf{y}\right)
    \label{eq:autopq_objective_advanced}
\end{equation}
is defined, with an arbitrary probabilistic performance metric as the objective function $Q$.
In contrast to \autopqdef, \autopqadv considers the point forecasting method's hyperparameter configuration $\boldsymbol{\uplambda}_\text{p}$ together with the sampling hyperparameter $\uplambda_\text{q}$ for generating samples from the point forecast's latent space representation.

In order to solve \eqref{eq:autopq_objective_advanced} efficiently, the computing times between training the point forecasting model and generating a quantile forecast based on this point forecast must be properly balanced.
\autoref{fig:autopq_computational-effort-total_bar-plot} compares these efforts across the six evaluation datasets.
Since generating a quantile forecast can be repeatedly refined with different sampling hyperparameters, we use two nested loops to solve \eqref{eq:autopq_objective_advanced} as outlined in \autoref{alg:autopq_combined-smart}.
The following paragraphs detail this joint optimization, the application of \ac{PK} during optimization, and the implementation details of the chosen algorithms.

\begin{algorithm}
    \caption{%
        The \autopq \acs{HPO} algorithm optimizes the hyperparameters of the point forecasting method and the sampling hyperparameter of the \acs{cINN} jointly, taking into account the significantly lower computational effort of the latter.
    }
    \label{alg:autopq_combined-smart}
    \begin{algorithmic}[1]
        \STATEx
        \REQUIRE
        $
        \mathbf{\Lambda}_\text{p},
        \mathbf{\Lambda}_\text{q},
        B_\text{i},
        B_\text{t},
        \mathbf{X}_\text{train},
        \mathbf{X}_\text{val},
        \mathbf{y}_\text{train},
        \mathbf{y}_\text{val},
        f_\text{\ac{cINN}}\left(\cdot\right)
        $
        \STATEx
        \STATEx\COMMENT{Initialize \texttt{Propulate}~\cite{Taubert2023} (\acs{EA})}
        \STATE $\texttt{trial\_generator\_p(}
                \mathbf{\Lambda}_\text{p}
            \texttt{)}$
        \STATE $b_\text{t} \leftarrow \SI{0}{\second}$
        \WHILE{$b_\text{t} < B_\text{t}$}
            \STATE $t \leftarrow
                \texttt{get\_current\_time()}$
            \STATE $\boldsymbol{\uplambda}_\text{p} \leftarrow
                \texttt{trial\_generator\_q.get\_config()}$
            \INDENT\COMMENT{High computational effort}
            \STATE $f_\text{p}\left(\cdot\right) \leftarrow
                \texttt{train\_p\_model(}
                    \boldsymbol{\uplambda}_\text{p},
                    \mathbf{X}_\text{train},
                    \mathbf{y}_\text{train}
                \texttt{)}$
            \INDENT\COMMENT{Distribution of $\uplambda_\text{q}^{\hat{\star}}$ in active population}
            \STATE $m, s \leftarrow \texttt{trial\_generator\_p.get\_stats()}$
            \INDENT\COMMENT{Initialize \texttt{Hyperopt}~\cite{Bergstra2013} (\acs{BO}) with \ac{PK}}
            \STATE $\texttt{trial\_generator\_q(}
                    \mathbf{\Lambda}_\text{q},
                    m,
                    s
                \texttt{)}$
            \FOR{$b_\text{i} \leftarrow 0$ \textbf{to} $B_\text{i}$}
                \STATE $\uplambda_\text{q} \leftarrow
                    \texttt{trial\_generator\_q.get\_config()}$
                \INDENT\COMMENT{Low computational effort}
                \STATE $f_\text{q}\left(\cdot\right) \leftarrow
                    \texttt{generate\_q\_model(}
                        f_\text{p}\left(\cdot\right),
                        f_\text{\ac{cINN}}\left(\cdot\right),
                        \uplambda_\text{q}
                    \texttt{)}$
                \INDENT\COMMENT{Low computational effort}
                \STATE $Q \leftarrow
                    \texttt{assess\_performance(}
                        f_\text{q}\left(
                            \mathbf{X}_\text{val},
                            \mathbf{y}_\text{val}
                        \right)
                \texttt{)}$
                \STATE $\texttt{trial\_generator\_q.update(}
                        Q,
                        \uplambda_\text{q}
                    \texttt{)}$
                \STATE $b_\text{i} \leftarrow b_\text{i} + 1$
                \IF {($\texttt{early\_stopping(}Q\texttt{)}$)}
                    \STATE \textbf{break}
                \ENDIF
            \ENDFOR
            \STATE $\uplambda_\text{q}^{\hat{\star}} \leftarrow
                \texttt{trial\_generator\_q.get\_best\_config()}$
            \INDENT\COMMENT{Store $\uplambda_\text{q}^{\hat{\star}}$ associated with $\boldsymbol{\uplambda}_\text{p}$}
            \STATE $\texttt{trial\_generator\_p.update(}
                Q,
                \boldsymbol{\uplambda}_\text{p},
                \uplambda_\text{q}^{\hat{\star}}
            \texttt{)}$
            \STATE $b_\text{t} \leftarrow
                b_\text{t} + \left(\texttt{get\_current\_time()} - t\right)$
        \ENDWHILE
        \STATE $\boldsymbol{\uplambda}_\text{p}^{\hat{\star}}, \uplambda_\text{q}^{\hat{\star}} \leftarrow
            \texttt{trial\_generator\_p.get\_best\_config()}$
        \STATEx
        \ENSURE
        $\boldsymbol{\uplambda}_\text{p}^{\hat{\star}}, \uplambda_\text{q}^{\hat{\star}}$
    \end{algorithmic}
    \vspace{4pt}\tiny
    $\mathbf{\Lambda}_\text{p}, \mathbf{\Lambda}_\text{q}$: configuration space (point forecast, quantile forecast); $B_{\text{i}}, B_{\text{t}}$: budget (iteration, time); $\mathbf{X}_\text{train}, \mathbf{X}_\text{val}$: model inputs (training, validation); $\mathbf{y}_\text{train}, \mathbf{y}_\text{val}$: model outputs (training, validation); $f_\text{\ac{cINN}}, f_\text{p}, f_\text{q}$: trained model (\ac{cINN}, point forecast, quantile forecast); $\boldsymbol{\uplambda}_\text{p}^{\hat{\star}}, \uplambda_\text{q}^{\hat{\star}}$: optimal configuration (point forecast, quantile forecast)
\end{algorithm}

\paragraph{Joint optimization.}
\autoref{alg:autopq_combined-smart} illustrates the nested joint \ac{HPO} applied in \autopq. 
The outer loop aims to identify the optimal hyperparameter configuration for the point forecasting method $\boldsymbol{\uplambda}_\text{p}^{\hat{\star}}$.
First, the outer loop trial generator suggests a candidate $\boldsymbol{\uplambda}_\text{p}$.
Importantly, after the random trial generator initialization, this suggestion is based on the probabilistic forecasting performance $Q$ of previous trials and not on their point forecasting performance.
Then, the point forecasting model $f_\text{p}\left(\cdot\right)$ is trained.
Afterward, the inner loop optimizes the corresponding sampling hyperparameter $\uplambda_\text{q}^{\hat{\star}}$ by solving \eqref{eq:autopq_objective_default}.
Specifically, the inner loop trial generator proposes a candidate $\uplambda_\text{q}$, and the quantile forecast $f_\text{q}\left(\cdot\right)$ is generated by passing the point forecast of $f_\text{p}\left(\cdot\right)$ forward through the \ac{cINN}, generating samples from the point forecast's latent space representation, and subsequently passing them backward through the \ac{cINN}.
Then, the probabilistic forecasting performance $Q$ is assessed, and the inner loop generator is updated with $Q$ and $\uplambda_\text{q}$.
The inner loop continues until an early stopping condition is met or the iteration budget $B_\text{i}$ is exhausted, yielding the optimal sampling hyperparameter $\uplambda_\text{q}^{\hat{\star}}$ associated with $\boldsymbol{\uplambda}_\text{p}$.
Before starting a new outer loop iteration, the outer trial generator is updated with the probabilistic forecasting performance $Q$, $\boldsymbol{\uplambda}_\text{p}$, and $\uplambda_\text{q}^{\hat{\star}}$ to suggest a new candidate $\boldsymbol{\uplambda}_\text{p}$ for the point forecasting model.
It continues until the time budget $B_\text{t}$ is exhausted, ultimately yielding the optimal configuration, \ie, $\boldsymbol{\uplambda}_\text{p}^{\hat{\star}}$ and the associated $\uplambda_\text{q}^{\hat{\star}}$.\footnote{
    The time budget refers to creating a model for one dataset.
}

\paragraph{Prior knowledge.}
In initializing the inner trial generator, we leverage \ac{PK} by assuming that the optimal sampling hyperparameter $\uplambda_\text{q}^{\hat{\star}}$ for a given point forecasting method configuration $\boldsymbol{\uplambda}_\text{p}$ is related to previously assessed configurations.
Consequently, the inner trial generator begins with statistical parameters derived from these past configurations.\footnote{
    These values are stored during outer trial generator updates.
}

\paragraph{Optimization algorithms.}
Since we do not have a closed\-/form expression of the configuration space $\mathbf{\Lambda}$ in \eqref{eq:autopq_objective_advanced} and the objective function $Q$ is noisy, non\-/differentiable, and potentially non\-/convex, solving \eqref{eq:autopq_objective_advanced} requires black\-/box optimization methods.
These methods serve as trial generators, suggesting new candidate configurations to be evaluated in both the inner and outer loop of \autoref{alg:autopq_combined-smart}.

Training the point forecasting model in the \textbf{outer loop} is computationally expensive and benefits from \ac{GPU} acceleration.
Using multiple \acp{GPU} enables evaluating multiple hyperparameter configurations $\boldsymbol{\uplambda}_\text{p}$ (\ie, multiple trials of $Q$) in parallel.
To handle varying training durations and avoid \ac{GPU} idle time, asynchronous optimization is applied.
For this purpose, we employ the Python package \texttt{Propulate}~\cite{Taubert2023} as the outer trial generator.
This evolutionary optimizer maintains a continuous population for asynchronous configuration assessments\footnote{
    The \ac{EA}'s setup is based on~\cite{Taubert2023} and detailed in \autoref{tab:autopq_propulate-configuration} in the Appendix.
}, which also enables incorporating \ac{PK} into the inner loop.

\begin{figure}[tb!]
    \centering
    \begin{subfigure}[t]{0.49\textwidth}
        \hfill
        \includegraphics{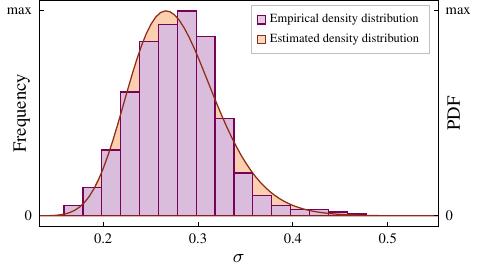}
        \vspace{-5mm}
        \caption{
            Estimated \acs{BO} prior distribution based on the optimal sampling hyperparameters associated and stored with the active and already assessed candidates of the \acs{EA} population, visualized as empirical density distribution.
            We assume a log\-/normal distribution and estimate its expected value $m$ and standard deviation $s$.
        }
        \label{fig:autopq_prior-distribution-estimation_mixed-plot}
    \end{subfigure}%
    \\[5mm]
    \begin{subfigure}[t]{0.49\textwidth}
        \includegraphics{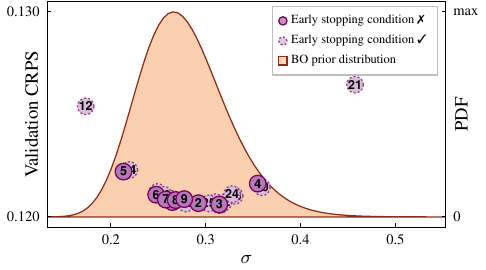}
        \vspace{-5mm}
        \caption{
            Initializing the \acs{BO} with the estimated prior distribution already provides a good estimate for the first trials, before the \acs{TPE} surrogate model fitted on these trials gains influence.
            To illustrate the progression after the early stopping condition is met at the ninth iteration, the \acs{BO} was continued until the \nth{25} iteration (dashed dots), with each dot representing an iteration.
            After reaching this condition, sampling also continues at the edges of the distribution (iterations 12 and 21).
        }
        \label{fig:autopq_prior-distribution-sampling_mixed-plot}
    \end{subfigure}%
    \caption{
        Estimated \acs{BO} prior distribution and its utilization in the \acs{HPO} of the sampling hyperparameter.
    }
    \label{fig:autopq_prior-distribution_estimation-sampling_multi-plot}
\end{figure}

Generating a probabilistic forecast based on a point forecast in the \textbf{inner loop} is computationally cheap.
Additionally, leveraging \ac{PK} enables rapid convergence, often resulting in early stopping.
Hence, parallelizing the inner loop would add unnecessary overhead without reducing computation time.
Instead, the inner loop runs sequentially using the Python package \texttt{Hyperopt}~\cite{Bergstra2013} as trial generator.
\texttt{Hyperopt} employs \ac{BO} with the \ac{TPE} as the surrogate model to explore and exploit the configuration space by estimating the distribution of well\-/performing hyperparameter configurations in relation to underperforming ones~\cite{Bischl2021}.
To initialize \texttt{Hyperopt}, we use \ac{PK} obtained from the outer loop trial generator \texttt{Propulate}.
Specifically, the mean $m$ and standard deviation $s$ of the optimal sampling hyperparameters' natural logarithms are computed from already assessed candidates in \texttt{Propulate}'s population (see \autoref{fig:autopq_prior-distribution-estimation_mixed-plot}). 
These values are then used to define \texttt{Hyperopt}'s prior distribution as $\log-\mathcal{N}\left(m, s\right)$.
Assuming a log\-/normal distribution excludes negative values and is valid, as demonstrated in \autoref{fig:autopq_prior-distribution-estimation_mixed-plot} with the \ac{MLP} on the Load-BW dataset.
A random set of $\uplambda_\text{q}$ values is initially drawn from this prior distribution to explore regions with high prior probabilities before fitting the \ac{TPE} surrogate model.
Afterward, the fitted \ac{TPE} is used to schedule new candidates of $\{\uplambda_{\text{q},0},\uplambda_{\text{q},1},\ldots\}$ through an acquisition function that balances exploration and exploitation.
Each assessed $\uplambda_{\text{q},i}$ (\ie, each trial of $Q$) updates the \ac{TPE} surrogate model with its performance $Q$, reducing the prior distribution's influence over iterations.
Thus, values in regions with lower prior probabilities are also considered, unless early stopping occurs beforehand, as demonstrated in \autoref{fig:autopq_prior-distribution-sampling_mixed-plot} with the \ac{MLP} on the Load-BW dataset.

\paragraph{Early stopping.}
As with \autopqdef, optimization is stopped when the objective function's value $Q$ has converged or reached a satisfactory value.
The inner loop is terminated if $Q$ reaches a plateau across trials, \ie, the standard deviation of the five best\-/performing trials is less than \num{0.0005} with a patience of five trials.

\paragraph{Forecasting method selection.}

\autopqadv aims to select the best point forecasting method while jointly optimizing its hyperparameters $\boldsymbol{\uplambda}_\text{p}$ and the corresponding sampling hyperparameter $\uplambda_\text{q}$.
To efficiently combine method selection and joint \ac{HPO}, we use successive halving pruning.
This method prunes underperforming configurations and re\-/allocates resources to more promising ones.
Starting from the considered candidate forecasting methods, successive halving (\autoref{alg:concept_successive-halving}) iteratively eliminates the worst\-/performing methods and re\-/allocates resources to the better\-/performing ones, until one top candidate remains when the total time budget $B_\text{t}$ is exhausted.
Crucially, \ac{HPO} for the remaining candidates continues from checkpoints of the previous pruning round, allowing optimization runs to be split over multiple \ac{HPC} jobs.

\begin{algorithm}[t!]
    \caption{%
        The successive halving pruning strategy re\-/allocates computing resources from unpromising areas in the \acs{CASH} configuration space to more promising ones.
    }
    \label{alg:concept_successive-halving}
    \begin{algorithmic}[1]
        \STATEx
        \REQUIRE
        $
        \mathbf{\Lambda},
        B_{\text{t}},
        \mathbf{X}_\text{train},
        \mathbf{X}_\text{val},
        \mathbf{y}_\text{train},
        \mathbf{y}_\text{val}
        $
        \STATEx
        \STATEx \COMMENT{Initialize result store}
        \STATE $\texttt{result\_store(}
            \mathbf{\Lambda}
            \texttt{)}$
        \WHILE{$\texttt{len(}\mathbf{\Lambda}\texttt{)} > 1$}
            \FOR{$\mathbf{\Lambda}_{i} \texttt{ in } \mathbf{\Lambda}$}
                \INDENTAFTERFOR\COMMENT{Parallel evaluation, split computation budget}
                \STATE $B_{\text{t},i}
                    \leftarrow B_{\text{t}} / \texttt{len(}
                        \mathbf{\Lambda}
                    \texttt{)}$
                \INDENT\COMMENT{Initialize \acs{HPO} algorithm with stored results}
                \STATE $\texttt{hpo\_algo(result\_store)}$
                \INDENT\COMMENT{Run until budget is exhausted}
                \STATE $\texttt{hpo\_algo.run(}
                    \mathbf{\Lambda}_{i},
                    \mathbf{X}_\text{train},
                    \mathbf{X}_\text{val},
                    \mathbf{y}_\text{train},
                    \mathbf{y}_\text{val},
                    B_{\text{t},i}
                    \texttt{)}$
                \STATE $Q^{\hat{\star}}, \boldsymbol{\uplambda}_{i}^{\hat{\star}}
                    \leftarrow \texttt{hpo\_algo.get\_best\_config(}
                        \mathbf{\Lambda}_i
                    \texttt{)}$
                \INDENT\COMMENT{Update result store}
                \STATE $\texttt{result\_store.update(}
                    Q_{i}^{\hat{\star}},
                    \boldsymbol{\uplambda}_{i}^{\hat{\star}}
                    \texttt{)}$
            \ENDFOR
            \INDENT\COMMENT{Shrink configuration space}
            \STATE $\mathbf{\Lambda} \leftarrow \texttt{result\_store.get\_best\_half()}$    
        \ENDWHILE
        \STATEx
        \ENSURE
        $\boldsymbol{\uplambda}^{\hat{\star}}$
    \end{algorithmic}
    \vspace{4pt}\tiny
    $\mathbf{\Lambda}$: configuration space; $B_{\text{t}}$: time budget; $\mathbf{X}_\text{train}, \mathbf{X}_\text{val}$: model inputs (training, validation); $\mathbf{y}_\text{train}, \mathbf{y}_\text{val}$: model outputs (training, validation); $\boldsymbol{\uplambda}^{\hat{\star}}$: optimal configuration
\end{algorithm}

\section{Evaluation}
\label{sec:evaluation}

To evaluate \autopq\footnote{
    A Python implementation of AutoPQ can be found on GitHub: \url{https://github.com/SMEISEN/AutoPQ}
}
comprehensively, we first assess its probabilistic forecasting performance with default and advanced configurations to analyze the impacts of \ac{HPO} on improving the probabilistic performance.
Second, we conduct an ablation study to evaluate the effectiveness of \autopqadv, which jointly optimizes the hyperparameters of the point forecaster and the \ac{cINN}'s sampling hyperparameter.
Third, we compare the computational efforts of \autopqdef and \autopqadv.

\subsection{Benchmarking}
\label{ssec:autopq_benchmarking}
The subsequent section outlines the experimental setup for assessing probabilistic forecasting performance, followed by a presentation of the results and key insights gained from the analysis.
\subsubsection{Experimental setup}

In the following, we detail the data utilized, the evaluation strategy employed, and the benchmarks considered.

\paragraph{Data.}

\autopq is evaluated on six distinct datasets: Load-BW, Load-\acs{GCP}, Mobility, Price, \ac{PV}, and \ac{WP}.
Each dataset is briefly described below, followed by details on pre\-/processing and feature selection.
Additional information is provided in \autoref{tab:autopq_data} in the Appendix.

The \textbf{Load-BW} dataset~\cite{Wiese2019} includes the gross electrical consumption of Baden-Württemberg, Germany, obtained from the \ac{OPSD} platform.
The regional electrical consumption reflects the aggregation of a large number of \acfp{GCP}.

The \textbf{Load-\ac{GCP}} dataset~\cite{Trindade2015} contains the electrical consumption time series of 370 \acp{GCP} within a distribution grid in Portugal.
The time series \texttt{MT\_158} is used to generate local forecasts, representing the local load of a single \ac{GCP}.

The \textbf{Mobility} dataset~\cite{Fanaee2013} comprises hourly records of bike rentals in Washington D.C., USA, which serve as an indicator of individual mobility within the smart grid context.
Relevant weather and seasonal information such as air temperature, wind speed, and whether a day is a workday, weekend or holiday, are included.

The \textbf{Price} dataset features zonal electricity prices from an unspecified location, provided in the \ac{GEFCom} 2014~\cite{Hong2016} price forecasting track.
This challenge involved 14 tasks requiring day\-/ahead forecasts based on exogenous time series data, including total electrical consumption and zonal price.
All 14 tasks are combined into one dataset for the evaluation.

The \textbf{\acs{PV}} dataset includes the power generation of an Australian \acf{PV} plant, sourced from the \ac{GEFCom} 2014~\cite{Hong2016} \ac{PV} power forecasting track.
This track comprised 16 day\-/ahead forecasting tasks, supplemented by \acp{NWP} for the site from \ac{ECMWF}, including \ac{GHI} and cloud cover.
For the evaluation, all 16 tasks are combined in one dataset.

The \textbf{\acs{WP}} dataset contains the power generation of an Australian wind farm, taken from the \ac{GEFCom} 2014~\cite{Hong2016} \acf{WP} forecasting track.
This track consisted of 16 day\-/ahead forecasting tasks, also with \acp{NWP} for the site from \ac{ECMWF} available, including wind speed and direction at a height of $\SI{10}{\meter}$ and $\SI{100}{\meter}$.
All 16 tasks are combined into one dataset for the evaluation.

Originally, all datasets considered are hourly resolved, except for the Load-\ac{GCP} dataset, which was resampled to hourly resolution.

\paragraph{Evaluation strategy.}
We evaluate the day\-/ahead forecasting performance of \autopq on the six datasets described above, comparing both \autopqdef and \autopqadv configurations.
To ensure comparability across the datasets, we normalize each dataset before creating separate training, validation, and test sub\-/sets.\footnote{
    For the forecasting methods \ac{MLP}, \ac{DeepAR}, \ac{NHiTS}, and \ac{TFT}, $\SI{20}{\percent}$ of the training dataset are hold\-/out for early stopping, \ie, the training process is terminated when the loss on the hold\-/out data increases.
}
Detailed sample indices for these splits and selected exogenous features are provided in \autoref{tab:autopq_data} in the Appendix.

We consider nine different point forecasting methods, choosing three from each method family: \ac{sARIMAX}, \ac{ETS}, and \ac{TBATS} from \ac{SM}; \ac{MLP}, \ac{SVR}, and \ac{XGB} from \ac{ML}; and \ac{DeepAR}, \ac{NHiTS}, and \ac{TFT} from \ac{DL}.
For \autopqdef, we train all nine methods with their default configurations $\boldsymbol{\uplambda}_\text{p,default}$, generate probabilistic forecasts from the point forecasts, and optimize the sampling hyperparameter $\uplambda_\text{q}$ for each model, where the best\-/performing one is selected afterward.\footnote{
    The results of \autopqdef for the datasets Load\-/\ac{GCP}, Mobility, Price, and \ac{PV} are taken from~\cite{Phipps2023a}, while the results for the additional datasets Load\-/BW and \ac{WP} are not yet published.
}
These results are compared to our evaluation of \autopqadv, which jointly optimizes both configurations, \ie, $\boldsymbol{\uplambda}_\text{p}$ and $\uplambda_\text{q}$, and applies \ac{CASH} with the successive halving pruning strategy for the specified time budget of $B_\text{t} = \SI{8}{\hour}$.

Two crucial properties in evaluating probabilistic forecasts are sharpness and calibration.
According to Gneiting et al.~\cite{Gneiting2007}, probabilistic forecasts should aim to maximize sharpness without negatively affecting the calibration.
To balance these properties, the \ac{CRPS}
\begin{equation}
    \text{CRPS} = \frac{1}{K} \sum^{K}_{k = 1} \int_{\mathbb{R}}\left(\hat{F}_\mathrm{Y}[k]\left(x\right) - \vmathbb{1}\{y\left[k\right] \leq x\}\right)^2 \mathrm{d}x,
    \label{eq:intro_crps}
\end{equation}
is used as metric for both tuning and assessment, averaged over all time points $K$.
In \eqref{eq:intro_crps}, $\hat{F}[k]$ is the estimated \ac{CDF} of the forecast at time point $k$ and the realized value $y[k]$ is translated into a degenerate distribution with the indicator function $\vmathbb{1}\{y\left[k\right] \leq x\}$, which is one if $y[k]$ is less than the integration variable $x$, and zero otherwise.

The evaluation is performed five times on each dataset, with the arithmetic mean and the standard deviation reported to account for stochastic effects in training and optimization.
We ensure consistent results using the same hardware for all runs.\footnote{
    Each run is performed with four parallel trials using two Intel Xeon Platinum 8368 \acp{CPU} with 76 cores, four NVIDIA A100-40 \acp{GPU} with $\SI{40}{\giga\byte}$ memory (\acp{GPU} are utilized only for \ac{DL}\-/based forecasting methods), and $\SI{256}{\giga\byte}$ \ac{RAM}, all provided by the \ac{HPC} system HoreKa.
}
Additionally, we report computing time and electricity consumption to quantify the computational effort and enable comparisons related to sustainability, as detailed in~\cite{Debus2023}.

\paragraph{Benchmarks.}
We compare the probabilistic forecasting performance of \autopq against multiple benchmarks, categorized into two groups: direct probabilistic forecasts and probabilistic forecasts derived from existing point forecasts.\footnote{
    The benchmark results for the datasets Load\-/\ac{GCP}, Mobility, Price, and \ac{PV} are taken from~\cite{Phipps2023a}, while the results for the additional datasets Load\-/BW and \ac{WP} are not yet published.
}

The first category includes the direct probabilistic forecasts \ac{DeepAR}, \acp{QRNN}, and the \ac{NNQF}, which are detailed in \autoref{sec:related_work}.
We implement \ac{DeepAR}~\cite{Salinas2020} using the Python package \texttt{PyTorch Forecasting}~\cite{Beitner2020}, the \acp{QRNN} with the Python package \texttt{Keras}~\cite{Chollet2015} using the built\-/in \ac{PL} function, and the \ac{NNQF}~\cite{GonzalezOrdiano2020} using the \ac{MLP} of the \texttt{Scikit-learn}~\cite{Pedregosa2011} Python package.

The second category includes the point forecast\-/based probabilistic forecasts Gaussian \acp{PI}, Empirical \acp{PI}, and Conformal \acp{PI}.
While all of these methods consider the residuals $r\left[k\right] = \left|\hat{y}\left[k\right] - y\left[k\right]\right|, \forall k \in \mathbb{N}_1^{K_\text{val}}$ between the point forecasts $\hat{y}\left[k\right]$ and the realized values $y\left[k\right]$ from a validation dataset, they differ in their approach to calculate \acp{PI}.
Gaussian \acp{PI}~\cite{Hyndman2021} assume Gaussian\-/distributed residuals and estimate the standard deviation $\sigma$ to calculate \acp{PI} centered around the point forecast, adjusting $\sigma$ by a factor corresponding to the desired confidence level.
In contrast, Empirical \acp{PI}~\cite{Williams1971} do not assume a specific parametric \ac{PDF} but use the empirical \ac{PDF} of residuals (\ie, sorting $r\left[k\right]$ from smallest to largest) to calculate \acp{PI} based on the desired confidence level.
Conformal \acp{PI} for multi\-/step\-/ahead forecasts~\cite{Stankeviciute2021} calibrate the \acp{PI} for each forecast horizon without assuming a specific parametric \ac{PDF}.
A critical non\-/conformity score is calculated for each $r\left[k\right]$, and the temporal dependencies between these scores across the forecast horizon are obtained using Bonferroni correction.
These corrected scores are used to calculate \acp{PI} centered around the point forecast.

For each point forecast\-/based probabilistic benchmark, we compare against the best\-/performing base forecast.
That is, we select the point forecasting method achieving the lowest \ac{CRPS} after calculating \acp{PI} using the Gaussian, the Empirical, and the Conformal approach, respectively.

\subsubsection{Results}
In the following, we summarize and visualize our benchmarking results, followed by an interpretation and discussion in \autoref{sec:discussion}.


\begin{table*}[t!]
    \centering
    \caption{
        The \acs{CRPS} evaluated on the hold\-/out test sets with methods categorized into three groups: i) direct probabilistic benchmark methods, ii) point forecast\-/based probabilistic benchmark methods, and iii) \autopq with the default and the advanced configuration.
        The results of the benchmarks and \autopqdef for the datasets Load-\acs{GCP}, Mobility, Price, and \acs{PV} originate from~\cite{Phipps2024}.
    }   
    \label{tab:autopq_benchmarking}
    \begin{adjustbox}{max width=\linewidth}
\begin{tabular}{r|ccc|ccc|cc|}
\cmidrule[\heavyrulewidth]{2-9}
& \textbf{\acs{DeepAR}} & \textbf{\acsp{QRNN}} & \textbf{\acs{NNQF}} & \textbf{Gaussian} & \textbf{Empirical} & \textbf{Conformal} & \textbf{AutoPQ} & \textbf{AutoPQ} \\
      &       &       &       & \textbf{\acsp{PI}} & \textbf{\acsp{PI}} & \textbf{\acsp{PI}} & \textbf{default} & \textbf{advanced} \\
\cmidrule[\heavyrulewidth]{2-9}
Load-BW & 0.192 & 0.145 & 0.208 & 0.156$\,\overset{\phantom{\boldsymbol{*}}}{\text{\fontsize{6}{6}\selectfont(1)}}$ & 0.143$\,\overset{\phantom{\boldsymbol{*}}}{\text{\fontsize{6}{6}\selectfont(1)}}$ & 0.143$\,\overset{\phantom{\boldsymbol{*}}}{\text{\fontsize{6}{6}\selectfont(1)}}$ & 0.147$\,\overset{\phantom{\boldsymbol{*}}}{\text{\fontsize{6}{6}\selectfont(1)}}$ & \textbf{0.138}$\,\overset{\phantom{\boldsymbol{*}}}{\text{\fontsize{6}{6}\selectfont(1)}}$ \\[-4pt]
      & $\scriptscriptstyle\pm$\scriptsize0.008 & $\scriptscriptstyle\pm$\scriptsize0.002 & $\scriptscriptstyle\pm$\scriptsize0.004 & $\scriptscriptstyle\pm$\scriptsize0.000 & $\scriptscriptstyle\pm$\scriptsize0.000 & $\scriptscriptstyle\pm$\scriptsize0.000 & $\scriptscriptstyle\pm$\scriptsize0.001 & $\scriptscriptstyle\pm$\scriptsize0.001 \\
Load-\acs{GCP} & 0.312 & 0.287 & 0.263 & 0.299$\,\overset{\phantom{\boldsymbol{*}}}{\text{\fontsize{6}{6}\selectfont(1)}}$ & 0.234$\,\overset{\phantom{\boldsymbol{*}}}{\text{\fontsize{6}{6}\selectfont(1)}}$ & 0.234$\,\overset{\phantom{\boldsymbol{*}}}{\text{\fontsize{6}{6}\selectfont(1)}}$ & 0.234$\,\overset{\phantom{\boldsymbol{*}}}{\text{\fontsize{6}{6}\selectfont(1)}}$ & \textbf{0.218}$\,\overset{\phantom{\boldsymbol{*}}}{\text{\fontsize{6}{6}\selectfont(1)}}$ \\[-4pt]
      & $\scriptscriptstyle\pm$\scriptsize0.009 & $\scriptscriptstyle\pm$\scriptsize0.002 & $\scriptscriptstyle\pm$\scriptsize0.001 & $\scriptscriptstyle\pm$\scriptsize0.000 & $\scriptscriptstyle\pm$\scriptsize0.000 & $\scriptscriptstyle\pm$\scriptsize0.000 & $\scriptscriptstyle\pm$\scriptsize0.002 & $\scriptscriptstyle\pm$\scriptsize0.001 \\
Mobility & 0.299 & 0.443 & 0.542 & 0.360$\,\overset{\phantom{\boldsymbol{*}}}{\text{\fontsize{6}{6}\selectfont(2)}}$ & 0.268$\,\overset{\phantom{\boldsymbol{*}}}{\text{\fontsize{6}{6}\selectfont(2)}}$ & 0.268$\,\overset{\phantom{\boldsymbol{*}}}{\text{\fontsize{6}{6}\selectfont(2)}}$ & 0.263$\,\overset{\phantom{\boldsymbol{*}}}{\text{\fontsize{6}{6}\selectfont(2)}}$ & \textbf{0.258}$\,\overset{\phantom{\boldsymbol{*}}}{\text{\fontsize{6}{6}\selectfont(2)}}$ \\[-4pt]
      & $\scriptscriptstyle\pm$\scriptsize0.007 & $\scriptscriptstyle\pm$\scriptsize0.006 & $\scriptscriptstyle\pm$\scriptsize0.004 & $\scriptscriptstyle\pm$\scriptsize0.021 & $\scriptscriptstyle\pm$\scriptsize0.015 & $\scriptscriptstyle\pm$\scriptsize0.015 & $\scriptscriptstyle\pm$\scriptsize0.004 & $\scriptscriptstyle\pm$\scriptsize0.005 \\
Price & 0.158 & 0.157 & 0.183 & 0.279$\,\overset{\phantom{\boldsymbol{*}}}{\text{\fontsize{6}{6}\selectfont(2)}}$ & 0.161$\,\overset{\phantom{\boldsymbol{*}}}{\text{\fontsize{6}{6}\selectfont(2)}}$ & 0.161$\,\overset{\phantom{\boldsymbol{*}}}{\text{\fontsize{6}{6}\selectfont(2)}}$ & 0.140$\,\overset{\phantom{\boldsymbol{*}}}{\text{\fontsize{6}{6}\selectfont(2)}}$  & \textbf{0.131}$\,\overset{\phantom{\boldsymbol{*}}}{\text{\fontsize{6}{6}\selectfont(3)}}$ \\[-4pt]
      & $\scriptscriptstyle\pm$\scriptsize0.005 & $\scriptscriptstyle\pm$\scriptsize0.002 & $\scriptscriptstyle\pm$\scriptsize0.003 & $\scriptscriptstyle\pm$\scriptsize0.008 & $\scriptscriptstyle\pm$\scriptsize0.005 & $\scriptscriptstyle\pm$\scriptsize0.005 & $\scriptscriptstyle\pm$\scriptsize0.006 & $\scriptscriptstyle\pm$\scriptsize0.001 \\
\acs{PV} & 0.151 & \textbf{0.101} & 0.119 & 0.208$\,\overset{\phantom{\boldsymbol{*}}}{\text{\fontsize{6}{6}\selectfont(1)}}$ & 0.125$\,\overset{\phantom{\boldsymbol{*}}}{\text{\fontsize{6}{6}\selectfont(1)}}$ & 0.125$\,\overset{\phantom{\boldsymbol{*}}}{\text{\fontsize{6}{6}\selectfont(1)}}$ & 0.106$\,\overset{\phantom{\boldsymbol{*}}}{\text{\fontsize{6}{6}\selectfont(1)}}$ & \textbf{0.101}$\,\overset{\phantom{\boldsymbol{*}}}{\text{\fontsize{6}{6}\selectfont(1)}}$ \\[-4pt]
      & $\scriptscriptstyle\pm$\scriptsize0.013 & $\scriptscriptstyle\pm$\scriptsize0.001 & $\scriptscriptstyle\pm$\scriptsize0.000 & $\scriptscriptstyle\pm$\scriptsize0.000 & $\scriptscriptstyle\pm$\scriptsize0.000 & $\scriptscriptstyle\pm$\scriptsize0.000 & $\scriptscriptstyle\pm$\scriptsize0.001 & $\scriptscriptstyle\pm$\scriptsize0.001 \\
\acs{WP}  & 0.618 & 0.374 & 0.394 & 0.419$\,\overset{\phantom{\boldsymbol{*}}}{\text{\fontsize{6}{6}\selectfont(1)}}$ & 0.373$\,\overset{\phantom{\boldsymbol{*}}}{\text{\fontsize{6}{6}\selectfont(1)}}$ & 0.373$\,\overset{\phantom{\boldsymbol{*}}}{\text{\fontsize{6}{6}\selectfont(1)}}$ & 0.377$\,\overset{\phantom{\boldsymbol{*}}}{\text{\fontsize{6}{6}\selectfont(1)}}$ & \textbf{0.362}$\,\overset{\phantom{\boldsymbol{*}}}{\text{\fontsize{6}{6}\selectfont(1)}}$ \\[-4pt]
      & $\scriptscriptstyle\pm$\scriptsize0.028 & $\scriptscriptstyle\pm$\scriptsize0.002 & $\scriptscriptstyle\pm$\scriptsize0.004 & $\scriptscriptstyle\pm$\scriptsize0.000 & $\scriptscriptstyle\pm$\scriptsize0.000 & $\scriptscriptstyle\pm$\scriptsize0.000 & $\scriptscriptstyle\pm$\scriptsize0.001 & $\scriptscriptstyle\pm$\scriptsize0.001\\
\cmidrule[\heavyrulewidth]{2-9}
Median & 0.245 & 0.222 & 0.236 & 0.289 & 0.197 & 0.197 & 0.191 & \textbf{0.178} \\
Mean  & 0.288 & 0.251 & 0.285 & 0.287 & 0.217 & 0.217 & 0.211 & \textbf{0.201} \\
\end{tabular}%
    \end{adjustbox}
    \scriptsize\\[2mm]
    \hspace{82mm}best\-/performing base point forecasting method: (1) \acs{XGB}, (2) \acs{TFT}, (3) \acs{NHiTS}
\end{table*}

Three key observations can be made from the benchmarking results in \autoref{tab:autopq_benchmarking}:
First, and most importantly, \autopqadv shows a significant performance advantage over \autopqdef and other benchmarks.
Second, the performance variability across datasets is greater for the direct probabilistic forecasting methods \ac{DeepAR}, \acp{QRNN}, and \ac{NNQF} compared to the point forecast\-/based probabilistic forecasting methods Gaussian, Empirical, and Conformal \acp{PI} as well as both \autopqdef and \autopqadv.
Third, the average performance across datasets is lower for \ac{DeepAR}, \acp{QRNN}, \ac{NNQF}, and Gaussian \acp{PI} compared to Empirical \acp{PI}, Conformal \acp{PI}, \autopqdef, and \autopqadv.

\begin{table*}
    \centering
    \caption{
        The percentage improvement of \autopqadv over the comparison methods in terms of the \acs{CRPS} evaluated on the hold\-/out test data.
        \autopqadv uses \acs{HPO} to optimize both the configuration of the point forecasting method $\boldsymbol{\uplambda}_\text{p}$ and the \acs{cINN}'s sampling hyperparameter $\uplambda_\text{q}$, while \autopqdef only optimizes $\uplambda_\text{q}$.
    }
    \label{tab:autopq_benchmarking-advanced_improvement}
    \begin{adjustbox}{max width=\linewidth}
\begin{tabular}{r|ccc|ccc|c|}
\cmidrule[\heavyrulewidth]{2-8}
& \textbf{\acs{DeepAR}} & \textbf{\acsp{QRNN}} & \textbf{\acs{NNQF}} & \textbf{Gaussian} & \textbf{Empirical} & \textbf{Conformal} & \textbf{AutoPQ} \\
      &       &       &       & \textbf{\acsp{PI}} & \textbf{\acsp{PI}} & \textbf{\acsp{PI}} & \textbf{default} \\
\cmidrule[\heavyrulewidth]{2-8}
Load-BW
    & \SI{28.1}{\percent}$\,\overset{\boldsymbol{*}}{\phantom{\text{\fontsize{6}{6}\selectfont(1)}}}$
    & \phantom{0}\SI{4.8}{\percent}$\,\overset{\boldsymbol{*}}{\phantom{\text{\fontsize{6}{6}\selectfont(1)}}}$
    & \SI{33.7}{\percent}$\,\overset{\boldsymbol{*}}{\phantom{\text{\fontsize{6}{6}\selectfont(1)}}}$
    & \SI{11.5}{\percent}$\,\overset{\boldsymbol{*}}{\text{\fontsize{6}{6}\selectfont(1)}}$
    & \phantom{0}\SI{3.5}{\percent}$\,\overset{\boldsymbol{*}}{\text{\fontsize{6}{6}\selectfont(1)}}$
    & \phantom{0}\SI{3.5}{\percent}$\,\overset{\boldsymbol{*}}{\text{\fontsize{6}{6}\selectfont(1)}}$
    & \SI{6.1}{\percent}$\,\overset{\boldsymbol{*}}{\text{\fontsize{6}{6}\selectfont(1)}}$
    \\
Load-\acs{GCP}
    & \SI{30.1}{\percent}$\,\overset{\boldsymbol{*}}{\phantom{\text{\fontsize{6}{6}\selectfont(1)}}}$
    & \SI{24.0}{\percent}$\,\overset{\boldsymbol{*}}{\phantom{\text{\fontsize{6}{6}\selectfont(1)}}}$
    & \SI{17.1}{\percent}$\,\overset{\boldsymbol{*}}{\phantom{\text{\fontsize{6}{6}\selectfont(1)}}}$
    & \SI{27.1}{\percent}$\,\overset{\boldsymbol{*}}{\text{\fontsize{6}{6}\selectfont(1)}}$
    & \phantom{0}\SI{6.8}{\percent}$\,\overset{\boldsymbol{*}}{\text{\fontsize{6}{6}\selectfont(1)}}$
    & \phantom{0}\SI{6.8}{\percent}$\,\overset{\boldsymbol{*}}{\text{\fontsize{6}{6}\selectfont(1)}}$
    & \SI{6.8}{\percent}$\,\overset{\boldsymbol{*}}{\text{\fontsize{6}{6}\selectfont(1)}}$
    \\
Mobility
    & \SI{13.7}{\percent}$\,\overset{\boldsymbol{*}}{\phantom{\text{\fontsize{6}{6}\selectfont(1)}}}$
    & \SI{41.8}{\percent}$\,\overset{\boldsymbol{*}}{\phantom{\text{\fontsize{6}{6}\selectfont(1)}}}$
    & \SI{52.4}{\percent}$\,\overset{\boldsymbol{*}}{\phantom{\text{\fontsize{6}{6}\selectfont(1)}}}$
    & \SI{28.3}{\percent}$\,\overset{\boldsymbol{*}}{\text{\fontsize{6}{6}\selectfont(2)}}$
    & \phantom{0}\SI{3.7}{\percent}$\,\overset{\phantom{\boldsymbol{*}}}{\text{\fontsize{6}{6}\selectfont(2)}}$
    & \phantom{0}\SI{3.7}{\percent}$\,\overset{\phantom{\boldsymbol{*}}}{\text{\fontsize{6}{6}\selectfont(2)}}$
    & \SI{1.9}{\percent}$\,\overset{\phantom{\boldsymbol{*}}}{\text{\fontsize{6}{6}\selectfont(2)}}$
    \\
Price
    & \SI{17.1}{\percent}$\,\overset{\boldsymbol{*}}{\phantom{\text{\fontsize{6}{6}\selectfont(1)}}}$
    & \SI{16.6}{\percent}$\,\overset{\boldsymbol{*}}{\phantom{\text{\fontsize{6}{6}\selectfont(1)}}}$
    & \SI{28.4}{\percent}$\,\overset{\boldsymbol{*}}{\phantom{\text{\fontsize{6}{6}\selectfont(1)}}}$
    & \SI{53.0}{\percent}$\,\overset{\boldsymbol{*}}{\text{\fontsize{6}{6}\selectfont(2)}}$
    & \SI{18.6}{\percent}$\,\overset{\boldsymbol{*}}{\text{\fontsize{6}{6}\selectfont(2)}}$
    & \SI{18.6}{\percent}$\,\overset{\boldsymbol{*}}{\text{\fontsize{6}{6}\selectfont(2)}}$
    & \SI{6.4}{\percent}$\,\overset{\boldsymbol{*}}{\text{\fontsize{6}{6}\selectfont(2)}}$
    \\
\acs{PV}
    & \SI{33.1}{\percent}$\,\overset{\boldsymbol{*}}{\phantom{\text{\fontsize{6}{6}\selectfont(1)}}}$
    & \phantom{0}\SI{0.0}{\percent}\phantom{$\,\overset{\boldsymbol{*}}{\text{\fontsize{6}{6}\selectfont(2)}}$}
    & \SI{15.1}{\percent}$\,\overset{\boldsymbol{*}}{\phantom{\text{\fontsize{6}{6}\selectfont(1)}}}$
    & \SI{51.4}{\percent}$\,\overset{\boldsymbol{*}}{\text{\fontsize{6}{6}\selectfont(1)}}$
    & \SI{19.2}{\percent}$\,\overset{\boldsymbol{*}}{\text{\fontsize{6}{6}\selectfont(1)}}$
    & \SI{19.2}{\percent}$\,\overset{\boldsymbol{*}}{\text{\fontsize{6}{6}\selectfont(1)}}$
    & \SI{4.7}{\percent}$\,\overset{\boldsymbol{*}}{\text{\fontsize{6}{6}\selectfont(1)}}$
    \\
\acs{WP}
    & \SI{41.4}{\percent}$\,\overset{\boldsymbol{*}}{\phantom{\text{\fontsize{6}{6}\selectfont(1)}}}$
    & \phantom{0}\SI{3.2}{\percent}$\,\overset{\boldsymbol{*}}{\phantom{\text{\fontsize{6}{6}\selectfont(1)}}}$
    & \phantom{0}\SI{8.1}{\percent}$\,\overset{\boldsymbol{*}}{\phantom{\text{\fontsize{6}{6}\selectfont(1)}}}$
    & \SI{13.6}{\percent}$\,\overset{\boldsymbol{*}}{\text{\fontsize{6}{6}\selectfont(1)}}$
    & \phantom{0}\SI{2.9}{\percent}$\,\overset{\boldsymbol{*}}{\text{\fontsize{6}{6}\selectfont(1)}}$
    & \phantom{0}\SI{2.9}{\percent}$\,\overset{\boldsymbol{*}}{\text{\fontsize{6}{6}\selectfont(1)}}$
    & \SI{4.0}{\percent}$\,\overset{\boldsymbol{*}}{\text{\fontsize{6}{6}\selectfont(1)}}$
    \\
\cmidrule[\heavyrulewidth]{2-8}
Mean
    & \SI{27.3}{\percent}\phantom{$\,\overset{\boldsymbol{*}}{\text{\fontsize{6}{6}\selectfont(2)}}$}
    & \SI{15.1}{\percent}\phantom{$\,\overset{\boldsymbol{*}}{\text{\fontsize{6}{6}\selectfont(2)}}$}
    & \SI{25.8}{\percent}\phantom{$\,\overset{\boldsymbol{*}}{\text{\fontsize{6}{6}\selectfont(2)}}$}
    & \SI{30.8}{\percent}\phantom{$\,\overset{\boldsymbol{*}}{\text{\fontsize{6}{6}\selectfont(2)}}$}
    & \phantom{0}\SI{9.1}{\percent}\phantom{$\,\overset{\boldsymbol{*}}{\text{\fontsize{6}{6}\selectfont(2)}}$}
    & \phantom{0}\SI{9.1}{\percent}\phantom{$\,\overset{\boldsymbol{*}}{\text{\fontsize{6}{6}\selectfont(2)}}$}
    & \SI{5.0}{\percent}\phantom{$\,\overset{\boldsymbol{*}}{\text{\fontsize{6}{6}\selectfont(2)}}$}
    \\
\end{tabular}%

    \end{adjustbox}
    \scriptsize\\[1mm]
    \hspace{33.5mm}$_{}^{\boldsymbol{*}}$ improvement is significant (p\-/value < $0.05$); best\-/performing base point forecasting method: (1) \acs{XGB}, (2) \acs{TFT}
\end{table*}

\autoref{tab:autopq_benchmarking-advanced_improvement} shows the percentage improvement of \autopqadv relative to the six benchmarks and \autopqdef per dataset.
Improvements are highlighted with an asterisk if they are significant, \ie, the p\-/value of the one\-/tailed t\-/test is less than $0.05$.
The following insights can be drawn from the table: 
First, we observe significant improvements in 38 out of 42 tests (across seven methods and six datasets).
Second, \autopqadv shows significant improvement over each direct probabilistic benchmark, averaging at least $\SI{15.1}{\percent}$, and each point forecast\-/based probabilistic benchmark, averaging at least $\SI{9.1}{\percent}$.
Third, \autopqadv outperforms \autopqdef by $\SI{5.0}{\percent}$ on average.

\subsubsection{Insights}

\begin{figure*}[tb!]
    \centering
    \begin{subfigure}[t]{0.34\textwidth}
        \includegraphics{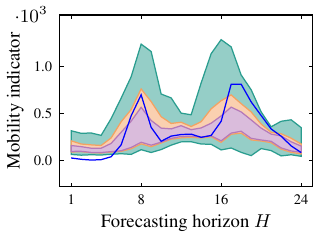}
        \caption{
            \autopq with \acs{ETS}.
        }
        \label{fig:autopq_prediction-intervals_CRPS_bike_ETS_1_mixed-plot}
    \end{subfigure}%
    \begin{subfigure}[t]{0.34\textwidth}
        \includegraphics{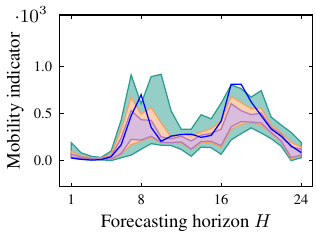}
        \caption{
            \autopq with \acs{XGB}.
        }
        \label{fig:autopq_prediction-intervals_CRPS_bike_XGB_6_mixed-plot}
    \end{subfigure}%
    \begin{subfigure}[t]{0.34\textwidth}
        \includegraphics{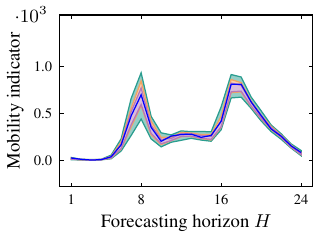}
        \caption{
            \autopq with \acs{TFT}.
        }
        \label{fig:autopq_prediction-intervals_CRPS_bike_TFT_1_mixed-plot}
    \end{subfigure}%
    \\[2mm]
    \begin{subfigure}[t]{0.34\textwidth}
        \includegraphics{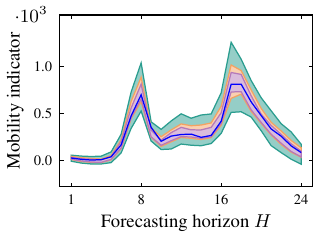}
        \caption{
            \acs{DeepAR} benchmark.
        }
        \label{fig:autopq_prediction-intervals_benchmarks_bike_DeepAR_1_mixed-plot}
    \end{subfigure}%
    \begin{subfigure}[t]{0.34\textwidth}
        \includegraphics{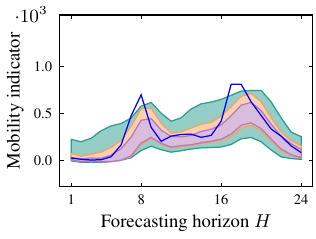}
        \caption{
            \acs{NNQF} benchmark.
        }
        \label{fig:autopq_prediction-intervals_benchmarks_bike_NNQF_1_mixed-plot}
    \end{subfigure}%
    \begin{subfigure}[t]{0.34\textwidth}
        \includegraphics{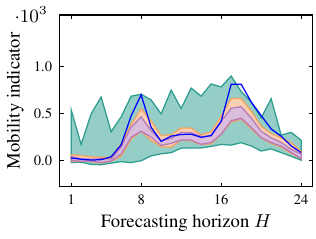}
        \caption{
            \acsp{QRNN} benchmark.
        }
        \label{fig:autopq_prediction-intervals_benchmarks_bike_QRNN_1_mixed-plot}
    \end{subfigure}%
    \\[2mm]
    \begin{subfigure}[t]{0.34\textwidth}
        \includegraphics{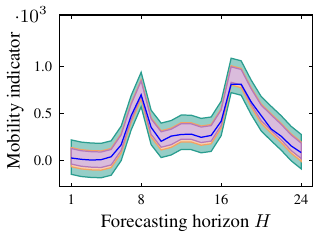}
        \caption{
            \acs{TFT} Gaussian \acsp{PI} benchmark.
        }
        \label{fig:autopq_prediction-intervals_benchmarks_bike_Gaussian-PI-TFT_1_mixed-plot}
    \end{subfigure}%
    \begin{subfigure}[t]{0.34\textwidth}
        \includegraphics{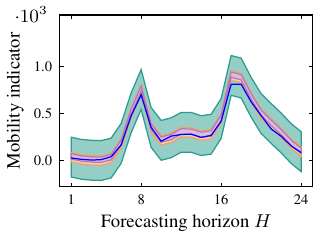}
        \caption{
            \acs{TFT} Empirical \acsp{PI} benchmark.
        }
        \label{fig:autopq_prediction-intervals_benchmarks_bike_Empirical-PI-TFT_1_mixed-plot}
    \end{subfigure}%
    \begin{subfigure}[t]{0.34\textwidth}
        \includegraphics{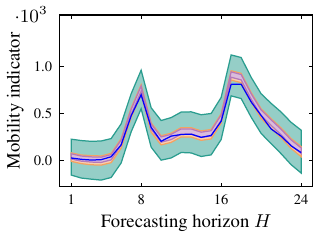}
        \caption{
            \acs{TFT} Conformal \acsp{PI} benchmark.
            }
        \label{fig:autopq_prediction-intervals_benchmarks_bike_Conformal-PI-TFT_1_mixed-plot}
    \end{subfigure}%
    \\[2mm]
    \begin{subfigure}[b]{1\textwidth}
        \centering
        \includegraphics{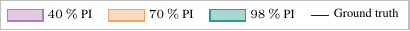}
    \end{subfigure}%
    \caption{
        Exemplary $\SI{40}{\percent}$, $\SI{70}{\percent}$, and $\SI{98}{\percent}$ \acsp{PI} for the Mobility dataset.
        The probabilistic forecasts of \autopqadv (a-c) are generated based on the point forecasting methods \acs{ETS} (\acs{SM}), \acs{XGB} (\acs{ML}), and \acs{TFT} (\acs{DL}), respectively, showing different performance.
        Note that \autopqadv automatically optimizes the hyperparameters and selects the best\-/performing method for each dataset, \ie, \ac{TFT} for Mobility.
        The probabilistic benchmarks (d-i) can be categorized into direct probabilistic methods (\acs{DeepAR}, \acs{NNQF}, \acsp{QRNN}) and point forecast\-/based probabilistic methods (Gaussian \acsp{PI}, Empirical \acsp{PI}, Conformal \acsp{PI} based on a \acs{TFT} forecaster).
    }
    \label{fig:autopq_prediction-intervals_CRPS_bike_multi-plot}
\end{figure*}

\autoref{fig:autopq_prediction-intervals_CRPS_bike_multi-plot} provides a visual comparison of \acp{PI} generated by \autopqadv using three exemplarily selected point forecasting methods (\ac{ETS}, \ac{XGB}, and \ac{TFT}) compared to the probabilistic benchmarks for the Mobility dataset.
The \autopqadv \acp{PI} are generated for the best\-/performing configuration identified through \ac{CASH} with successive halving, with \ac{TFT} being the best method for this dataset (as detailed later in \autoref{fig:autopq_successive-halving_evaluation_multi-plot}).
\autoref{fig:autopq_prediction-intervals_CRPS_bike_multi-plot} (a-c) shows that the \acp{PI} generated from the three point forecasting methods vary in sharpness, as reflected in their widths.
Notably, clear differences are evident between the considered methods \ac{TFT}, \ac{XGB}, and \ac{ETS}, ranking first, fourth, and ninth in the \ac{CASH}, respectively.

Another observation concerns the \acp{PI} when the realized values are near zero or zero.
Since the forecasted value (mobility indicator) is restricted to non\-/negative values, negative \acp{PI} are unreasonable.
\autopq's \acp{PI} account for this restriction, unlike the probabilistic benchmarks' \acp{PI}, cf. \autoref{fig:autopq_prediction-intervals_CRPS_bike_multi-plot} (a-c) and (d-i).
However, the post\-/processing required for \autopq's quantiles to comply with this non\-/zero constraint on the \ac{CRPS} is minimal, as detailed in \autoref{tab:autopq_post-processing} in the Appendix.

\subsection{Ablation study}
This section outlines the experimental setup of the ablation study, which evaluates the effectiveness of the automation methods applied in \autopqadv, followed by the presentation of results and insights.

\subsubsection{Experimental setup}

In the following, we outline the data used, the evaluation strategy employed, and the ablations considered in our experimental setup.

\paragraph{Data.}
As for benchmarking, the ablation study of \autopqadv is performed on six different datasets: Load-BW, Load-\ac{GCP}, Mobility, Price, \ac{PV}, and \ac{WP} (see \autoref{ssec:autopq_benchmarking} for an overview).

\paragraph{Evaluation strategy.}
The ablation study is based on a similar evaluation strategy as the previous benchmarking.
Specifically, the six datasets outlined in \autoref{ssec:autopq_benchmarking} are normalized and divided into training, validation, and test sets, as detailed in \autoref{tab:autopq_data} in the Appendix.

To reduce the computational effort, we only consider the \ac{MLP} and \ac{NHiTS} point forecasting methods without using successive halving.
This selection is reasoned by i) the computational effort, ii) the probabilistic performance, and iii) the sensitivity to \ac{HPO}.
As illustrated in \autoref{fig:autopq_computational-effort-total_bar-plot}, \ac{MLP} and \ac{NHiTS} require low and moderate to high computational effort, respectively.
In addition, both methods achieve competitive probabilistic performance and respond well to \ac{HPO}, as shown later in \autoref{tab:autopq_successive-halfing_improvement}.

In line with the benchmarking, the \ac{CRPS} metric \eqref{eq:intro_crps} is used for both tuning and assessment, with evaluations conducted five times on each dataset.
To ensure consistent runtime measurements, we utilize the same hardware as for the benchmarking in all runs (see \autoref{ssec:autopq_benchmarking}).

\paragraph{Ablation 1.}
The effectiveness of the inner loop in \autoref{alg:autopq_combined-smart}, responsible for finding the optimal sampling hyperparameter $\uplambda_\text{q}^{\hat{\star}}$, is assessed.
For comparison, a conventional \ac{HPO} is used that does not take into account the different computational efforts required for training the point forecasting model and generating probabilistic forecasts.
This is achieved by omitting the inner loop of \autoref{alg:autopq_combined-smart}, resulting in \autoref{alg:autopq_combined-dumb}.
In this setup, the trial generator suggests configurations for both the point forecaster $\boldsymbol{\uplambda}_\text{p}$ and the sampling hyperparameter $\uplambda_\text{q}$ as combined configuration space $\boldsymbol{\uplambda} = \boldsymbol{\uplambda}_\text{p} \times \uplambda_\text{q}$.
Thus, only one sampling hyperparameter is assessed for each trained point forecasting model.
To ensure comparability, both \autoref{alg:autopq_combined-smart} and \ref{alg:autopq_combined-dumb} are run without successive halving, \ie, using the full time budget $B_\text{t} = \SI{8}{\hour}$ and considering the same configuration spaces.

\begin{algorithm}
    \caption{%
        \acs{HPO} algorithm for optimizing the point forecasting method's hyperparameters and the \acs{cINN}'s sampling hyperparameter simultaneously without considering the significantly lower computational effort of the latter.
    }
    \label{alg:autopq_combined-dumb}
    \begin{algorithmic}[1]
        \STATEx
        \REQUIRE
        $
        \mathbf{\Lambda}_\text{p},
        \mathbf{\Lambda}_\text{q},
        B_\text{i},
        B_\text{t},
        \mathbf{X}_\text{train},
        \mathbf{X}_\text{val},
        \mathbf{y}_\text{train},
        \mathbf{y}_\text{val},
        f_\text{\ac{cINN}}\left(\cdot\right)
        $        
        \STATEx
        \STATEx\COMMENT{Initialize \texttt{Propulate}~\cite{Taubert2023} (\acs{EA})}
        \STATE $\texttt{trial\_generator(}
                \mathbf{\Lambda}_\text{p}, \mathbf{\Lambda}_\text{q}
            \texttt{)}$
        \STATE $b_\text{t} \leftarrow \SI{0}{\second}$
        \WHILE{$b_\text{t} < B_\text{t}$}
            \STATE $t \leftarrow
                \texttt{get\_current\_time()}$
            \STATE $\boldsymbol{\uplambda}_\text{p}, \uplambda_\text{q} \leftarrow
                \texttt{trial\_generator.get\_config(}
                    \mathbf{\Lambda}_\text{p},
                    \mathbf{\Lambda}_\text{q}
                \texttt{)}$
            \INDENT\COMMENT{High computational effort}
            \STATE $ f_\text{p}\left(\cdot\right) \leftarrow
                \texttt{train\_p\_model(}
                    \boldsymbol{\uplambda}_\text{p},
                    \mathbf{X}_\text{train},
                    \mathbf{y}_\text{train}
                \texttt{)}$
            \INDENT\COMMENT{Low computational effort}
            \STATE $ f_\text{q}\left(\cdot\right) \leftarrow
                \texttt{generate\_q\_model(}
                     f_\text{p}\left(\cdot\right),
                    f_\text{\ac{cINN}}\left(\cdot\right),
                    \uplambda_\text{q}
                \texttt{)}$
            \INDENT\COMMENT{Low computational effort}
            \STATE $Q \leftarrow
                \texttt{assess\_performance}(
                     f_\text{q}\left(
                        \mathbf{X}_\text{val},
                        \mathbf{y}_\text{val}
                    \right)
                \texttt{)}$
            \STATE $\texttt{trial\_generator.update(}
                    \boldsymbol{\uplambda}_\text{p},
                    \uplambda_\text{q},
                    Q
                \texttt{)}$
            \STATE $b_\text{t} \leftarrow
                b_\text{t} + \left(\texttt{get\_current\_time()} - t\right)$
        \ENDWHILE
        \STATE $\boldsymbol{\uplambda}_\text{p}^{\hat{\star}}, \uplambda_\text{q}^{\hat{\star}} \leftarrow
            \texttt{trial\_generator.get\_best\_config(}
                \mathbf{\Lambda}_\text{p},
                \mathbf{\Lambda}_\text{q}
            \texttt{)}$
        \STATEx
        \ENSURE
        $\boldsymbol{\uplambda}_\text{p}^{\hat{\star}}, \uplambda_\text{q}^{\hat{\star}}$
    \end{algorithmic}
    \vspace{4pt}\tiny
    $\mathbf{\Lambda}_\text{p}, \mathbf{\Lambda}_\text{q}$: configuration space (point forecast, quantile forecast); $B_{\text{i}}, B_{\text{t}}$: budget (iteration, time); $\mathbf{X}_\text{train}, \mathbf{X}_\text{val}$: model inputs (training, validation); $\mathbf{y}_\text{train}, \mathbf{y}_\text{val}$: model outputs (training, validation); $f_\text{\ac{cINN}}, f_\text{p}, f_\text{q}$: trained model (\ac{cINN}, point forecast, quantile forecast); $\boldsymbol{\uplambda}_\text{p}^{\hat{\star}}, \uplambda_\text{q}^{\hat{\star}}$: optimal configuration (point forecast, quantile forecast)
\end{algorithm}

\paragraph{Ablation 2.}
Different trial generators for the inner loop of \autoref{alg:autopq_combined-smart} are compared.
Specifically, we compare \ac{PK}\-/based \ac{BO} against random search and \ac{BO}, neither of which uses \ac{PK}.
The comparison is based on the number of iterations required to fulfill the early stopping condition outlined in \autoref{sssec:autopq_advanced}.
To avoid the computationally expensive task of repeating the entire \ac{CASH} process for all six datasets five times across all nine forecasting methods, we limit the comparison to repeating only the inner loop of \autoref{alg:autopq_combined-smart}.
Thus, the inner loop is executed five times for each dataset using the best configuration for each forecasting method and evaluated with the two trial generators being compared.

\paragraph{Ablation 3.}
The decision quality of the successive halving is evaluated, specifically the impact of pruning underperforming forecasting methods in the \ac{CASH}.
To achieve this, we compare the evolution of the forecasting methods' validation performances with successive halving to their performance evolution without pruning.
Each pruned forecasting method from the benchmarking (see \autoref{ssec:autopq_benchmarking}) is continued until the full time budget $B_\text{t} = \SI{8}{\hour}$ is exhausted.
This allows us to analyze whether successive halving retains forecasting methods that ultimately perform well while pruning those that do not improve with additional computational effort.
Additionally, we evaluate how the nine forecasting methods respond to \ac{HPO}, \ie, the improvement over the point forecasting methods' default hyperparameter configurations.

\subsubsection{Results}
The following section summarizes the results of the ablation study and provides supporting visualizations, which are further interpreted and discussed in \autoref{sec:discussion}.

\begin{figure}[hp!]
    \centering
    \begin{subfigure}[t]{0.51\textwidth}
        \includegraphics{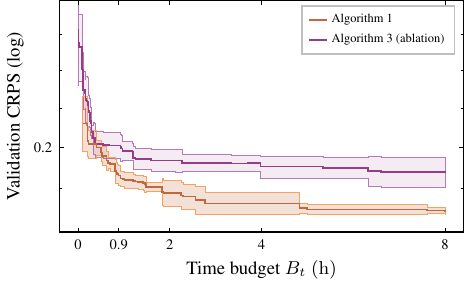}
        \caption{
            Convergence of \acs{MLP} on Price.
        }
        \label{fig:autopq_convergence-runs_price_MLP_line-plot}
    \end{subfigure}%
    \\[4mm]
    \begin{subfigure}[t]{0.51\textwidth}
        \includegraphics{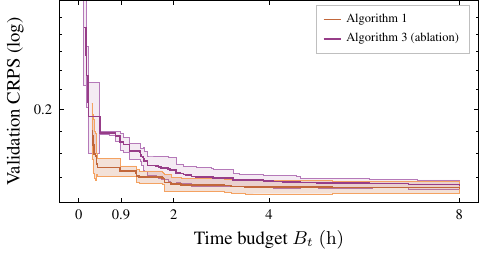}
        \caption{
            Convergence of \acs{NHiTS} on Price.
        }
        \label{fig:autopq_convergence-runs_price_NHiTS_line-plot}
    \end{subfigure}%
    \caption{
        Comparison of the convergence of \autoref{alg:autopq_combined-smart} and \ref{alg:autopq_combined-dumb} (ablation) for the \acs{HPO} of the \acs{MLP} and \acs{NHiTS} on the two exemplary datasets.
        The thick solid line represents the mean value, and the opaque area is the standard deviation over five runs.
    }
    \label{fig:autopq_convergence-runs_multi-plot}
\end{figure}

\paragraph{Ablation 1.}
The results comparing the effectiveness of the originally proposed \autopq \acs{HPO} \autoref{alg:autopq_combined-smart} to its ablated version \autoref{alg:autopq_combined-dumb} are presented in \autoref{fig:autopq_convergence-runs_multi-plot} for the \ac{MLP} and \ac{NHiTS}.
These figures exemplarily illustrate the \ac{CRPS} evolution on the validation dataset over the \ac{HPO} budget.
The time budget marks for the pruning rounds are also shown to contextualize the validation performance within successive halving.
We can draw two key observations from these figures:
First, the mean value of \autoref{alg:autopq_combined-smart} is consistently lower than that of \autoref{alg:autopq_combined-dumb} for most of the \ac{HPO} time budget.
Second, the mean value of \autoref{alg:autopq_combined-smart} is notably lower than that of \autoref{alg:autopq_combined-dumb} at the beginning of the \ac{HPO}.
This gap narrows toward the end of the \ac{HPO} time budget, with both values appearing to converge to the same value.\footnote{
    Both observations also hold for the other four datasets and both of the considered forecasting methods.
}

\begin{figure}[tb!]
    \centering
    \includegraphics{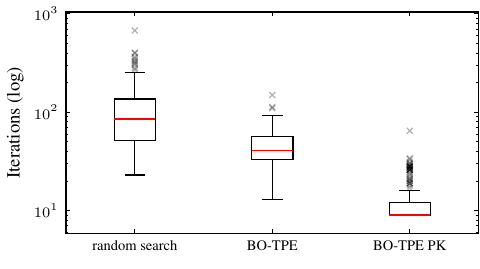}
    \caption{
        Number of iterations required for the inner loop of \autoref{alg:autopq_combined-smart} until the early stopping criterion is fulfilled, compared for three trial generators: random search, \acs{BO}-\acs{TPE}, and \acs{BO}-\acs{TPE}-\acs{PK}.
        The box plot considers all nine forecasting methods, the six datasets, and five runs, \ie, 270 data tuples per trial generator.
    }
    \label{fig:autopq_n-iter_box-plot}
\end{figure}

\paragraph{Ablation 2.}
\autoref{fig:autopq_n-iter_box-plot} compares the results of different trial generators for the inner loop in \autoref{alg:autopq_combined-smart}.
The figure shows the number of iterations required until the early stopping condition is met, aggregated over all nine forecasting methods, the six datasets, and five runs.\footnote{
    The computational effort scales approximately linearly with the number of iterations since the initialization effort is negligible.
}
Three key observations emerge:
First, \ac{BO} using the \ac{TPE} surrogate model significantly reduces the number of iterations compared to random search.
Second, integrating \ac{PK} into \ac{BO}-\ac{TPE} further decreases the number of required iterations.
Notably, even the upper outliers fall within the range of the lower whisker of \ac{BO}-\ac{TPE} without \ac{PK}.
Third, all three trial generators achieve similar validation \ac{CRPS}, as detailed in \autoref{tab:autopq_n-iter_score} in the Appendix.
Specifically, \ac{BO}-\ac{TPE}-\ac{PK} delivers similar probabilistic performance while requiring significantly fewer iterations compared to the other two generators.

\begin{figure}[tb!]
    \centering
    \begin{subfigure}[t]{1\columnwidth}
        \includegraphics{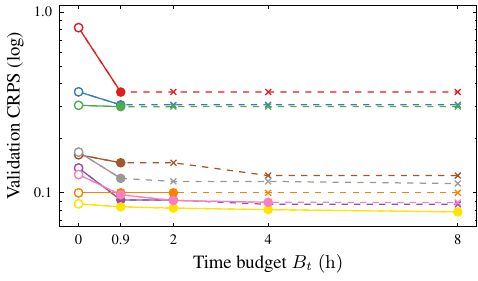}
        \caption{
            Successive halving pruning rounds on Load-BW.
        }
        \label{fig:autopq_successive-halving_opsd_1_line-plot}
    \end{subfigure}%
    \\[2mm]
    \begin{subfigure}[t]{1\columnwidth}
        \includegraphics{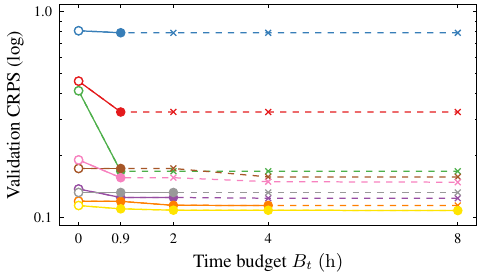}
        \caption{
            Successive halving pruning rounds on \acs{PV}.
        }
        \label{fig:autopq_successive-halving_solar_1_line-plot}
    \end{subfigure}%
    \\[2mm]
        \includegraphics{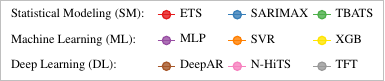}
    \caption{
        Evolution of the considered forecasting methods' validation performances in the successive halving\-/based \acs{CASH} of \autopqadv.
        The performance at $\SI{0}{\hour}$ reflects \autopqdef, shown as a reference and thus not included in the time budget $B_\text{t}$.
        Solid lines with dot markers show the performance of active methods, while dotted lines with cross markers show the performance of inactive methods as it would have evolved without pruning.
        For simplicity, the progress in the successive halving pruning rounds is assumed to be linear.
    }
    \label{fig:autopq_successive-halving_evaluation_multi-plot}
\end{figure}

\paragraph{Ablation 3.}
\autoref{fig:autopq_successive-halving_evaluation_multi-plot} shows the results of the successive halving ablation study.\footnote{
    \autoref{fig:autopq_successive-halving_evaluation_multi-plot} shows the representative results of run no. 1.
}
Three key findings are notable:
First, the best\-/performing forecasting method remains consistent across all datasets, meaning it is not erroneously deactivated during pruning.
Second, this best\-/performing method is already identified by \autopqdef.
Rank changes occur among the following methods during successive halving.
Third, while \ac{SM} methods show significant improvements in the first pruning round (\autopqadv) compared to their default configuration (\autopqdef), they still rank last even with additional time budget.
In contrast, the middle and upper ranks are competitive among methods from the \ac{ML} and \ac{DL} families.

\subsubsection{Insights}

\begin{figure}[tb!]
    \centering
    \begin{subfigure}[t]{1\columnwidth}
        \includegraphics{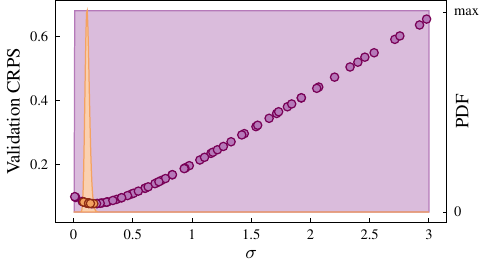}
        \vspace{-5mm}
        \caption{
            $\uplambda_\text{q}$-\acs{HPO} on Load-BW.
        }
        \label{fig:autopq_prior-distribution_opsd_xgb_1_mixed-plot}
    \end{subfigure}%
    \\[2mm]
    \begin{subfigure}[t]{1\columnwidth}
        \includegraphics{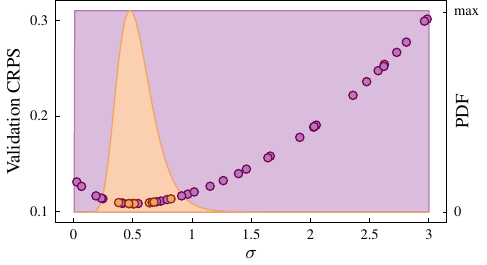}
        \vspace{-5mm}
        \caption{
            $\uplambda_\text{q}$-\acs{HPO} on \acs{PV}.
        }
        \label{fig:autopq_prior-distribution_solar_xgb_1_mixed-plot}
    \end{subfigure}%
    \\[2mm]
        \includegraphics{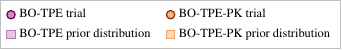}
    \caption{
        The \acs{HPO} of the sampling hyperparameter conducted in the inner loop of \autoref{alg:autopq_combined-smart} using \acs{BO}-\acs{TPE} and \acs{BO}-\acs{TPE}-\acs{PK}.
        The validation performance (\acs{CRPS}) over the trial configurations ($\uplambda_\text{q}$) is represented as dots, and the \acs{TPE} prior distribution is shown as a filled area normalized by the maximum of the respective \acs{PDF}.
        The plots illustrate the \acs{HPO} for a single \acs{XGB} configuration but are representative of the other point forecasting methods.
    }
    \label{fig:autopq_prior-distribution_multi-plot}
\end{figure}

\autoref{fig:autopq_prior-distribution_multi-plot} illustrates the impact of leveraging \ac{PK} for \ac{BO}-\ac{TPE} on each dataset.
The plots depict the \ac{CRPS} on the validation dataset together with the prior distribution over the sampling hyperparameter $\sigma$ for both \ac{BO}-\ac{TPE} with and without \ac{PK}.
For \ac{BO}-\ac{TPE}, a continuous uniform \ac{PDF} $\mathcal{U}(0,3)$ is assumed, while \ac{BO}-\ac{TPE}-\ac{PK} uses a log\-/normal \ac{PDF} $\mathcal{N}(\mu, \sigma)$ with the parameters derived from the \ac{EA} population.
The \acp{PDF} in \autoref{fig:autopq_prior-distribution_multi-plot} are normalized to their maximum value to ensure visual comparability.
These plots, while only exemplarily showing the sampling hyperparameter optimization for the \ac{XGB} point forecast, are representative of other point forecasting methods concerning the following observations:
First, the optimal sampling hyperparameter $\uplambda_\text{q}^{\hat{\star}} = \sigma^{\hat{\star}}$ is highly dependent on the dataset, showing differences in both location and sensitivity.
For instance, the validation \ac{CRPS} valley in \autoref{fig:autopq_prior-distribution_solar_xgb_1_mixed-plot} is flatter than the one in \autoref{fig:autopq_prior-distribution_opsd_xgb_1_mixed-plot}.
Correspondingly, the prior distribution of \ac{BO}-\ac{TPE}-\ac{PK} shows varying spread.
Second, \autoref{fig:autopq_prior-distribution_multi-plot} confirms our earlier findings that
\ac{BO}-\ac{TPE}-\ac{PK} requires significantly fewer iterations to meet the early stopping condition.
This condition is often satisfied before the \ac{TPE} surrogate model starts trading off exploitation and exploration.

Examining the improvement in the validation \ac{CRPS} without pruning as shown in \autoref{fig:autopq_successive-halving_evaluation_multi-plot}, we can calculate the percentage improvement. 
Results are listed in \autoref{tab:autopq_successive-halfing_improvement}, where significant improvements with a p\-/value of the one\-/tailed t\-/test smaller than $0.05$ are highlighted with an asterisk.
Three observations stand out:
First, the \ac{SM} methods show inconsistent improvements;
some exhibit large and significant enhancements, while others demonstrate little to no improvement.
Second, the \ac{ML} methods achieve significant improvements, with the \ac{MLP} showing the best response to \ac{HPO}, improving on average by $\SI{12.3}{\percent}$.
Third, the \ac{DL} methods achieve significant improvements only for \ac{DeepAR} and \ac{NHiTS}, with average enhancements by $\SI{14.7}{\percent}$ and $\SI{15.8}{\percent}$, respectively.
In contrast, the \ac{TFT} exhibits a weak response to \ac{HPO}, showing no significant improvement in five of the six datasets.

\begin{table*}
    \centering
    \caption{
        The percentage improvement of each optimized configuration over the corresponding default configuration in the successive halving ablation study without pruning in terms of the \acs{CRPS} evaluated on the validation data.
    }
    \label{tab:autopq_successive-halfing_improvement}
    \begin{adjustbox}{max width=\linewidth}
\begin{tabular}{r|rrr|rrr|rrr|}
\cmidrule[\heavyrulewidth]{2-10}
& \multicolumn{1}{c}{\textbf{\acs{ETS}}\phantom{\textsuperscript{(3)}}} & \multicolumn{1}{c}{\textbf{\acs{sARIMAX}}\phantom{\textsuperscript{(3)}}} & \multicolumn{1}{c}{\textbf{\acs{TBATS}}\phantom{\textsuperscript{(3)}}} & \multicolumn{1}{c}{\textbf{\acs{MLP}}\phantom{\textsuperscript{(3)}}} & \multicolumn{1}{c}{\textbf{\acs{SVR}}\phantom{\textsuperscript{(3)}}} & \multicolumn{1}{c}{\textbf{\acs{XGB}}\phantom{\textsuperscript{(3)}}} & \multicolumn{1}{c}{\textbf{\acs{DeepAR}}\phantom{\textsuperscript{(3)}}} & \multicolumn{1}{c}{\textbf{\acs{NHiTS}}\phantom{\textsuperscript{(3)}}} & \multicolumn{1}{c}{\textbf{\acs{TFT}}\phantom{\textsuperscript{(3)}}} \\
\cmidrule[\heavyrulewidth]{2-10}
Load-BW
    & \SI{57.7}{\percent}$\,\overset{\boldsymbol{*}}{\phantom{\text{\fontsize{6}{6}\selectfont(3)}}}$
    & \SI{12.9}{\percent}$\,\overset{\boldsymbol{*}}{\phantom{\text{\fontsize{6}{6}\selectfont(3)}}}$
    & \SI{0.5}{\percent}$\,\overset{\phantom{\boldsymbol{*}}}{\phantom{\text{\fontsize{6}{6}\selectfont(3)}}}$
    & \SI{29.2}{\percent}$\,\overset{\boldsymbol{*}}{\text{\fontsize{6}{6}\selectfont(3)}}$
    & \SI{0.3}{\percent}$\,\overset{\phantom{\boldsymbol{*}}}{\phantom{\text{\fontsize{6}{6}\selectfont(3)}}}$
    & \SI{7.5}{\percent}$\,\overset{\boldsymbol{*}}{\text{\fontsize{6}{6}\selectfont(1)}}$
    & \SI{25.6}{\percent}$\,\overset{\boldsymbol{*}}{\phantom{\text{\fontsize{6}{6}\selectfont(3)}}}$
    & \SI{30.5}{\percent}$\,\overset{\boldsymbol{*}}{\text{\fontsize{6}{6}\selectfont(2)}}$
    & \SI{23.5}{\percent}$\,\overset{\boldsymbol{*}}{\phantom{\text{\fontsize{6}{6}\selectfont(3)}}}$
    \\
Load-\acs{GCP}
    & <\SI{0.1}{\percent}$\,\overset{\phantom{\boldsymbol{*}}}{\phantom{\text{\fontsize{6}{6}\selectfont(3)}}}$
    & <\SI{0.1}{\percent}$\,\overset{\phantom{\boldsymbol{*}}}{\phantom{\text{\fontsize{6}{6}\selectfont(3)}}}$
    & \SI{8.0}{\percent}$\,\overset{\phantom{\boldsymbol{*}}}{\phantom{\text{\fontsize{6}{6}\selectfont(3)}}}$
    & \SI{6.6}{\percent}$\,\overset{\boldsymbol{*}}{\phantom{\text{\fontsize{6}{6}\selectfont(3)}}}$
    & \SI{4.1}{\percent}$\,\overset{\boldsymbol{*}}{\text{\fontsize{6}{6}\selectfont(3)}}$
    & \SI{4.8}{\percent}$\,\overset{\boldsymbol{*}}{\text{\fontsize{6}{6}\selectfont(1)}}$
    & \SI{16.7}{\percent}$\,\overset{\boldsymbol{*}}{\phantom{\text{\fontsize{6}{6}\selectfont(3)}}}$
    & \SI{4.8}{\percent}$\,\overset{\boldsymbol{*}}{\phantom{\text{\fontsize{6}{6}\selectfont(3)}}}$
    & \SI{6.7}{\percent}$\,\overset{\phantom{\boldsymbol{*}}}{\text{\fontsize{6}{6}\selectfont(2)}}$
\\
Mobility
    & \SI{42.3}{\percent}$\,\overset{\boldsymbol{*}}{\phantom{\text{\fontsize{6}{6}\selectfont(3)}}}$
    & \SI{40.5}{\percent}$\,\overset{\boldsymbol{*}}{\phantom{\text{\fontsize{6}{6}\selectfont(3)}}}$
    & \SI{45.0}{\percent}$\,\overset{\boldsymbol{*}}{\phantom{\text{\fontsize{6}{6}\selectfont(3)}}}$
    & \SI{4.9}{\percent}$\,\overset{\boldsymbol{*}}{\phantom{\text{\fontsize{6}{6}\selectfont(3)}}}$
    & \SI{4.4}{\percent}$\,\overset{\boldsymbol{*}}{\phantom{\text{\fontsize{6}{6}\selectfont(3)}}}$
    & \SI{6.8}{\percent}$\,\overset{\boldsymbol{*}}{\phantom{\text{\fontsize{6}{6}\selectfont(3)}}}$
    & \SI{8.9}{\percent}$\,\overset{\boldsymbol{*}}{\text{\fontsize{6}{6}\selectfont(2)}}$
    & \SI{14.2}{\percent}$\,\overset{\boldsymbol{*}}{\text{\fontsize{6}{6}\selectfont(3)}}$
    & \SI{1.0}{\percent}$\,\overset{\phantom{\boldsymbol{*}}}{\text{\fontsize{6}{6}\selectfont(1)}}$
\\
Price
    & \SI{2.2}{\percent}$\,\overset{\phantom{\boldsymbol{*}}}{\phantom{\text{\fontsize{6}{6}\selectfont(3)}}}$
    & \SI{52.5}{\percent}$\,\overset{\boldsymbol{*}}{\phantom{\text{\fontsize{6}{6}\selectfont(3)}}}$
    & <\SI{0.1}{\percent}$\,\overset{\phantom{\boldsymbol{*}}}{\phantom{\text{\fontsize{6}{6}\selectfont(3)}}}$
    & \SI{8.9}{\percent}$\,\overset{\boldsymbol{*}}{\text{\fontsize{6}{6}\selectfont(2)}}$
    & \SI{2.3}{\percent}$\,\overset{\boldsymbol{*}}{\phantom{\text{\fontsize{6}{6}\selectfont(3)}}}$
    & \SI{2.5}{\percent}$\,\overset{\boldsymbol{*}}{\phantom{\text{\fontsize{6}{6}\selectfont(3)}}}$
    & \SI{7.9}{\percent}$\,\overset{\boldsymbol{*}}{\phantom{\text{\fontsize{6}{6}\selectfont(3)}}}$
    & \SI{13.7}{\percent}$\,\overset{\boldsymbol{*}}{\text{\fontsize{6}{6}\selectfont(1)}}$
    & \SI{5.3}{\percent}$\,\overset{\phantom{\boldsymbol{*}}}{\text{\fontsize{6}{6}\selectfont(3)}}$
\\
\acs{PV}
    & \SI{37.5}{\percent}$\,\overset{\boldsymbol{*}}{\phantom{\text{\fontsize{6}{6}\selectfont(3)}}}$
    & \SI{14.3}{\percent}$\,\overset{\phantom{\boldsymbol{*}}}{\phantom{\text{\fontsize{6}{6}\selectfont(3)}}}$
    & \SI{56.7}{\percent}$\,\overset{\boldsymbol{*}}{\phantom{\text{\fontsize{6}{6}\selectfont(3)}}}$
    & \SI{10.2}{\percent}$\,\overset{\boldsymbol{*}}{\text{\fontsize{6}{6}\selectfont(3)}}$
    & \SI{4.3}{\percent}$\,\overset{\boldsymbol{*}}{\text{\fontsize{6}{6}\selectfont(2)}}$
    & \SI{6.2}{\percent}$\,\overset{\boldsymbol{*}}{\text{\fontsize{6}{6}\selectfont(1)}}$
    & \SI{13.8}{\percent}$\,\overset{\boldsymbol{*}}{\phantom{\text{\fontsize{6}{6}\selectfont(3)}}}$
    & \SI{22.0}{\percent}$\,\overset{\boldsymbol{*}}{\phantom{\text{\fontsize{6}{6}\selectfont(3)}}}$
    & \SI{6.9}{\percent}$\,\overset{\phantom{\boldsymbol{*}}}{\phantom{\text{\fontsize{6}{6}\selectfont(3)}}}$
\\
\acs{WP}
    & <\SI{0.1}{\percent}$\,\overset{\phantom{\boldsymbol{*}}}{\phantom{\text{\fontsize{6}{6}\selectfont(3)}}}$
    & <\SI{0.1}{\percent}$\,\overset{\phantom{\boldsymbol{*}}}{\phantom{\text{\fontsize{6}{6}\selectfont(3)}}}$
    & <\SI{0.1}{\percent}$\,\overset{\phantom{\boldsymbol{*}}}{\phantom{\text{\fontsize{6}{6}\selectfont(3)}}}$
    & \SI{13.9}{\percent}$\,\overset{\boldsymbol{*}}{\text{\fontsize{6}{6}\selectfont(3)}}$
    & \SI{3.6}{\percent}$\,\overset{\boldsymbol{*}}{\text{\fontsize{6}{6}\selectfont(2)}}$
    & \SI{5.6}{\percent}$\,\overset{\boldsymbol{*}}{\text{\fontsize{6}{6}\selectfont(1)}}$
    & \SI{15.0}{\percent}$\,\overset{\boldsymbol{*}}{\phantom{\text{\fontsize{6}{6}\selectfont(3)}}}$
    & \SI{9.3}{\percent}$\,\overset{\boldsymbol{*}}{\phantom{\text{\fontsize{6}{6}\selectfont(3)}}}$
    & \SI{8.5}{\percent}$\,\overset{\phantom{\boldsymbol{*}}}{\phantom{\text{\fontsize{6}{6}\selectfont(3)}}}$
\\
\cmidrule[\heavyrulewidth]{2-10}
\end{tabular}%

    \end{adjustbox}
    \scriptsize\\
    \hspace{18mm}$_{}^{\boldsymbol{*}}$ improvement is significant (p\-/value < $0.05$); (1), (2), (3): rank number in successive halving
\end{table*}

\subsection{Electricity consumption-awareness}
Our benchmarking results demonstrate that \autopqadv can significantly enhance performance compared to \autopqdef.
Since this improvement incurs a higher computational effort, the following evaluation aims to increase the awareness for energy\-/intensive performance improvements.

\subsubsection{Experimental setup}

Our evaluation is based on the average computational effort across the six datasets used in the benchmarking (\autoref{ssec:autopq_benchmarking}).
We measure the computational effort in terms of the electricity consumption because it is independent of data\-/center\-/specific or geopolitical factors at the time of collection ~\cite{Debus2023}.\footnote{
    Other metrics based on electricity consumption like the carbon footprint can be calculated using assumed or local conditions.
}
Since the research\-/exclusive \ac{HPC} system HoreKa\footnote{
    \url{https://www.nhr.kit.edu/userdocs/horeka/}
} was used for benchmarking, we additionally estimate the monetary costs that would be incurred by the Amazon AWS cloud computing service with equivalent hardware.

\subsubsection{Results}
The following section summarizes the electricity consumption required for performance improvements and corresponding monetary costs, which are further interpreted and discussed in \autoref{sec:discussion}.

\paragraph{Electricity consumption.}

\begin{figure}[tb!]
    \centering
    \begin{subfigure}[t]{1\columnwidth}
        \includegraphics{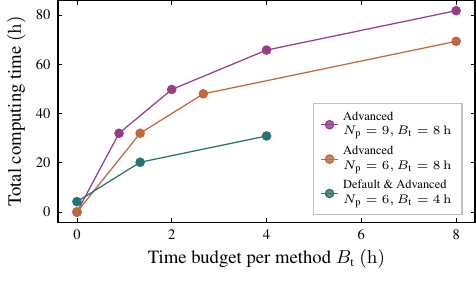}
        \vspace{-8mm}
        \caption{
            Evolution of the total computing time.
        }
        \label{fig:autopq_successive-halving-effort-variants_line-plot}
    \end{subfigure}%
    \\[2mm]
    \begin{subfigure}[t]{1\columnwidth}
        \includegraphics{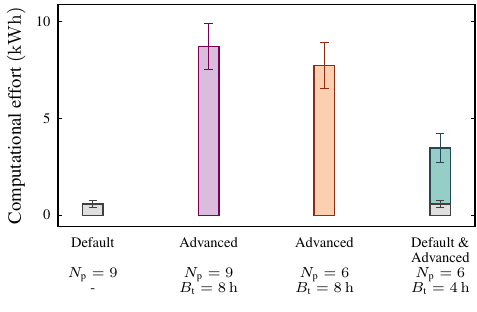}
        \vspace{-6mm}
        \caption{
            Resulting total electricity consumption.
        }
        \label{fig:autopq_kilo-watt-hours_bar-plot}
    \end{subfigure}%
    \caption{
        Comparison of the computational effort for a run of \autopqdef and \autopqadv for different numbers of point forecasting methods $N_\text{p}$ in the \acs{CASH} and different time budgets $B_\text{t}$ in the successive halving pruning strategy.
        Since \autopqadv assesses four trial configurations in parallel while \autopqdef runs sequentially, we multiply the total computing time of \autopqadv by four.
        The total electricity consumption was recorded by the HoreKa \ac{HPC} system.
    }
    \label{fig:autopq_computational-effort_multi-plot}
\end{figure}

\autoref{fig:autopq_computational-effort_multi-plot} shows the total computing time and the corresponding electricity consumption of \autopqdef and \autopqadv required for a single run, averaged across datasets and runs.

While a single \autopqdef run uses $\SI{4.22}{\hour}$ of compute time and consumes $\SI{0.57}{\kilo\watt{}\hour}$ of electrical energy, \autopqadv requires $\SI{81.78}{\hour}$ and consumes $\SI{8.73}{\kilo\watt{}\hour}$ for a run, with $N_\text{p}=9$ considered point forecasting methods and a time budget $B_\text{t}=\SI{8}{\hour}$.
This increased resource usage is due to the parallelized \ac{CASH}, which evaluates four trial configurations simultaneously for each point forecasting method. 
As a result, the total computing time accumulates over all pruning rounds of successive halving.
Specifically, the total computing time is calculated as the product of the number of active forecasting methods per round, the time budget per round, and the number of parallel trials conducted.

The benchmarking shows that the three \ac{SM} point forecasting methods perform significantly worse than the \ac{ML}- and \ac{DL}\-/based methods.
Omitting \ac{SM} methods due to their weak performance, \ie, $N_\text{p}=6$, would reduce the computational effort to $\SI{69.33}{\hour}$ and $\SI{7.75}{\kilo\watt{}\hour}$.

The ablation study reveals that the performance of \autopqdef, which uses default hyperparameters for the point forecasting methods $\boldsymbol{\uplambda}_\text{p,default}$ with an optimized sampling hyperparameter $\uplambda_\text{q}^{\hat{\star}}$, reflects the potential performance improvements achievable through \ac{HPO} to identify $\boldsymbol{\uplambda}_\text{p}^{\hat{\star}}$ with \autopqadv.
Thus, \autopqdef can be viewed as the initial pruning round in the successive halving\-/based \ac{CASH} of \autopqadv.
Since the performance gains in the final pruning round are minimal, the overall time budget could be further reduced to $B_\text{t}=\SI{4}{\hour}$.
Both adjustments would decrease the computational effort to $\SI{30.89}{\hour}$ and $\SI{3.47}{\kilo\watt{}\hour}$.

\paragraph{Monetary costs.}
We calculate the monetary costs of using Amazon AWS for both \autopqdef and \autopqadv, based on equivalent hardware on the Amazon EC2 G4 instances~\cite{AWS2024}.
For \autopqadv, the four\-/\ac{GPU} instance \texttt{g4ad.16xlarge} ($\SI{3.468}{\$\per\hour}$ on demand) closely matches the hardware utilized on the HoreKa \ac{HPC} system.
Since four trials are evaluated in parallel, the billed computing time shown in \autoref{fig:autopq_computational-effort_multi-plot} is reduced by a factor of four.
As the \ac{HPO} of the sampling hyperparameter is performed sequentially for \autopqdef, the single\-/\ac{GPU} instance \texttt{g4ad.4xlarge} ($\SI{0.867}{\$\per\hour}$ on demand) is comparable to the hardware utilized on HoreKa.
\begin{table*}
    \centering
    \caption{
        The monetary costs of using the Amazon EC2 G4 instances with hardware similar to the HoreKa \ac{HPC} system.
    }
    \label{tab:autopq_monetary-costs}
\begin{tabular}{lrrrr}
\toprule
\autopq configuration & Default & Advanced & Advanced & Default \& Advanced \\
Point forecasting methods & $N_\text{p}=9$ & $N_\text{p}=9$ & $N_\text{p}=6$ & $N_\text{p}=6$ \\
Successive halving budget & -     & $B_\text{t}=\SI{8}{\hour}$ & $B_\text{t}=\SI{4}{\hour}$ & $B_\text{t}=\SI{4}{\hour}$ \\
Amazon EC2 G4 instance & \texttt{g4ad.4xlarge} & \texttt{g4ad.16xlarge} & \texttt{g4ad.16xlarge} & \texttt{g4ad.16xlarge} \\
On-demand price & $\SI{0.87}{\$\per\hour}$ & $\SI{3.47}{\$\per\hour}$ & $\SI{3.47}{\$\per\hour}$ & $\SI{3.47}{\$\per\hour}$ \\
Computing hours & $\SI{4.22}{\hour}$ & $\SI{20.45}{\hour}$ & $\SI{17.33}{\hour}$ & $\SI{7.72}{\hour}$ \\
\midrule[\heavyrulewidth]
Monetary costs & $\SI{3.66}{\$}$ & $\SI{70.9}{\$}$ & $\SI{60.11}{\$}$ & $\SI{26.78}{\$}$ \\
\bottomrule
\end{tabular}%

\end{table*}

The costs for employing \autopqdef and \autopqadv on the above Amazon EC2 G4 instances are summarized in \autoref{tab:autopq_monetary-costs}.
Using \autopqdef would cost $\SI{3.66}{\$}$, while using \autopqadv would cost $\SI{26.78}{\$}$ with the electricity consumption\-/aware configuration ($N_\text{p}=6, B_\text{t}=\SI{4}{\hour}$).
For simplicity, we assume that all forecasting methods utilize \ac{GPU} instances, even though only the three \ac{DL} methods benefit from them.
To optimize costs and avoid \ac{GPU} idle time, \ac{CPU}\-/only instances should be employed for \ac{SM} and \ac{ML} point forecasting methods, as used for the benchmarking on HoreKa.

\section{Discussion}
\label{sec:discussion}
The subsequent section discusses the results of the benchmarking, the ablation studies, as well as limitations and benefits.

\subsection{Benchmarking}
We discuss two aspects of the benchmarking results for probabilistic day\-/ahead forecasting performance across the six datasets:
First, we observe a greater performance variability among the direct forecasting methods, \ie, \ac{DeepAR}, \acp{QRNN}, and \ac{NNQF}, compared to the point forecast\-/based probabilistic forecasting methods, \ie, Gaussian, empirical, and conformal \acp{PI}, as well as \autopqdef and \autopqadv.
The reduced variability in the latter group can be attributed to the probabilistic forecast being based on multiple point forecasting methods, from which the best-performing method is selected.
This performance\-/based selection ensures that only the top method prevails, thus mitigating the risk of poor performance associated with any method underperforming on the dataset.
Second, \autopqadv consistently outperforms \autopqdef and all benchmarking methods.
Specifically, the average improvement of \autopqadv surpasses that of \autopqdef when compared to both direct probabilistic benchmarks and point forecast\-/based probabilistic benchmarks.
This improvement is significant for five of the six datasets, suggesting that \autopqadv is particularly well\-/suited for performance\-/critical smart grid applications subject to high decision costs.


\subsection{Ablation study}
Our results of the ablation study reveal three key points:
First, the performance advantage of \autoref{alg:autopq_combined-smart} over \autoref{alg:autopq_combined-dumb}, evident at the beginning of the \ac{HPO}, is 
decisive for the effectiveness successive halving.
Since \autoref{alg:autopq_combined-dumb} (ablation) risks prematurely pruning forecasting methods that could potentially perform well, its performance deficit is particularly high for the initial and subsequent pruning rounds.
This is because the \ac{EA} requires several generations to identify a suitable sampling hyperparameter, with random initialization potentially yielding both suitable and unsuitable parameters.
In contrast, \autoref{alg:autopq_combined-smart} effectively identifies the optimal sampling hyperparameter for each trained model, thereby reducing this variability.
We observe this benefit for both \ac{NHiTS} (moderate to high computational effort) and the \ac{MLP} (low computational effort), demonstrating the general effectiveness of \autoref{alg:autopq_combined-smart} across different forecasting methods.
Second, leveraging \ac{PK} for the inner loop trial generator significantly reduces the number of iterations.
Maintaining a low iteration count is crucial for achieving the performance advantage of \autoref{alg:autopq_combined-smart} over \autoref{alg:autopq_combined-dumb} (ablation).
This improvement stems from the fact that the inner loop's effectiveness relies on keeping the computational effort needed to identify the optimal sampling hyperparameter low compared to the time required to train the model.
Third, successive halving proves effective in both pruning underperforming forecasting methods and retaining promising ones.
We find that the three \ac{SM} methods consistently perform poorly across all datasets and lag far behind \ac{ML} and \ac{DL} methods.
Therefore, we recommend to exclude these methods to reduce the computational effort.
For \ac{ML} and \ac{DL} methods, the performance of their default hyperparameter configuration already provides a strong indicator of the method’s potential for performance improvement in the \ac{HPO}.
Thus, if one method significantly outperforms another using default configurations, this performance gap is unlikely to change substantially through \ac{HPO}.

\subsection{Electricity consumption-awareness}
In the context of electricity consumption-awareness in view of sustainability, \autopq offers two configurations tailored to accommodate varying computing systems and performance needs: \autopqdef for general\-/purpose computing systems, delivering high\-/quality probabilistic forecasts, and \autopqadv, which necessitates \ac{HPC} systems to enhance forecasting quality further.
The latter uses a built\-/in successive halving strategy for optimizing resource use by pruning underperforming configurations and re\-/allocating computational resources to more promising ones.
We recommend running \autopqdef first and using the model if the forecasting quality is satisfactory.
Otherwise, the results can be used to initialize the successive halving\-/based \ac{CASH} of \autopqadv, which can improve the \ac{CRPS} up to $\SI{6.8}{\percent}$.
In terms of the efforts required for performance improvements, we quantify the energy footprint and monetary costs of both \autopqdef and \autopqadv.
Given that the monetary costs are relatively low compared to the personnel expenses associated with an iterative manual design process, we conclude that both \autopqdef and \autopqadv are cost\-/effective for most smart grid applications.
For applications in which decision\-/making heavily relies on forecasting quality, we anticipate that \autopqadv will quickly amortize its higher initial costs, ultimately becoming more cost\-/efficient than \autopqdef in the long run.

\subsection{Limitations}
\label{ssec:autopq_limitations}
With regard to the limitations of \autopq, two key aspects are discussed: possibly untapped potential for further performance improvements and desirable probabilistic properties.

\paragraph{Possibly untapped performance improvement potential.}
Performance could be further improved through the configuration space design, the feature selection, and integrating \ac{PK}.
An improperly sized configuration space may either waste computational resources on insensitive hyperparameters or overlook sensitive hyperparameters, missing potential performance gains.
While a comprehensive sensitivity analysis across various datasets is impractical due to high computational costs, \autopq's configuration space design is based on best practices.
Specifically, the hyperparameters and their value ranges are chosen according to recommendations provided in the respective forecasting methods' documentation.
The feature selection for the evaluation is based on~\cite{Phipps2024} (\autopqdef) to ensure comparability in benchmarking.
However, automated selection is preferable for real\-/world applications.
This limitation could be addressed by using automated feature selection methods, \eg, the \ac{MRMR} filter method~\cite{Ding2003}.

\paragraph{Desirable probabilistic properties.}
The benchmarking demonstrates that \autopq can be optimized to effectively minimize \ac{CRPS}.
Moreover, our previous work~\cite{Phipps2023a} highlights that optimizing the sampling hyperparameter for a customized metric allows the forecast to be tailored to different probabilistic properties.
However, it remains unclear which probabilistic properties should be prioritized for specific real\-/world smart grid applications.
This gap could be addressed by directly measuring the so\-/called forecast value of a configuration within the application, rather than relying on a probabilistic metric as a proxy.
For instance, the forecast value could be assessed by using the resulting costs in a forecast\-/based electricity cost optimization problem, as shown in~\cite{Werling2023}.\footnote{
    In this study, several \ac{MLP}\-/based point forecasting models are trained using different loss functions, and the model with the highest forecast value is selected.
}
This approach enables the \ac{HPO} to assess how well the resulting probabilistic properties align with the application's needs, rather than relying on a chosen probabilistic metric as a proxy.

\subsection{Benefits}
\autopq's ability to automatically design high\-/quality probabilistic forecasts with tailored properties is essential for scaling smart grid applications.
With the cost\-/effectiveness of training \autopqadv, we anticipate a swift return on investments for many smart grid applications, particularly where forecast quality is critical for decision\-/making.
A key advantage of our approach is its versatility: the \ac{cINN} can generate quantiles or \acp{PI} for any point forecasting method.
This flexibility enables the integration of new forecasting methods into \autopq's \ac{CASH} process, while consistently removing underperforming point forecasting methods.
Moreover, the joint optimization of the point forecasting methods' hyperparameters and the \ac{cINN}'s sampling hyperparameter is not limited to mathematically derivable validation metrics.
This is crucial because it supports the use of any metrics for the assessment in the \ac{HPO} process, including customized metrics or direct measurements of the so\-/called forecast value within smart grid applications, as previously suggested.

\section{Conclusion and outlook}
\label{sec:conclusion}
Designing probabilistic time series forecasting models for smart grid applications includes several challenges, \ie, i) quantifying forecast uncertainty in an unbiased and accurate manner, ii) automating the selection and enhancement of forecasting models to meet application needs, and iii) quantifying the environmental impacts required for performance improvements in view of sustainability.

In order to address these challenges, we introduce \autopq, a novel method for generating probabilistic forecasts.
\autopq uses a \ac{cINN} to convert point forecasts into quantile forecasts, leveraging existing well\-/designed point forecasting methods.
\autopq automates the selection and optimization of these methods, ensuring the best model is chosen and fine\-/tuned to improve probabilistic forecasting performance.
To handle varying performance needs and available computing power, \autopq comes with two configurations:
the default configuration suitable for providing high\-/quality probabilistic forecasts on general\-/purpose computing systems, and the advanced configuration for further enhancing forecast quality on \ac{HPC} systems.

The evaluation demonstrates that \autopq is both cost\-/effective and scalable, with significant performance improvements in forecasting quality over existing direct probabilistic forecasting benchmarks, averaging at least $\SI{15.1}{\percent}$, and point forecast\-/based probabilistic benchmarks, averaging at least $\SI{9.1}{\percent}$.
Additionally, the evaluation addresses environmental concerns by assessing \autopq's strategies for saving computing resources.
Specifically, the energy consumption can be reduced by $\SI{60}{\percent}$ by adjusting \autopqadv's successive halving strategy based on the ablation study results (omit \ac{SM} methods, reduce time budget, and use \autopqdef as initial pruning round).
Furthermore, information on \autopq's electricity consumption and monetary costs required for performance improvements is reported.
Specifically, the average improvement in forecast quality of $\SI{5}{\percent}$ of \autopqadv over \autopqdef additionally consumes $\SI{2.9}{\kilo\watt{}\hour}$ and additionally costs $\SI{23.12}{\$}$.

Future work will focus on refining the feature selection of \autopq, and incorporating the forecast value as the assessment metric to tailor the forecast directly to the application's needs, rather than using a probabilistic metric like the \ac{CRPS} as a proxy.

\section*{Acknowledgments}
This project is funded by the Helmholtz Association under the Program ``Energy System Design'', the Helmholtz Association's Initiative and Networking Fund through Helmholtz AI, and was performed on the HoreKa supercomputer funded by the Ministry of Science, Research and the Arts Baden-Württemberg and by the Federal Ministry of Education and Research.

Conceptualization and methodology:      S. M., K. P.;
Literature review:                      S. M., K. P.;
Data curation:                          K. P.;
Formal analysis:                        S. M.;
Structure:                              S. M.;
Visualization and layout:               S. M.;
Writing -- original draft preparation:  S. M.;
Writing -- review and editing:          S. M., K. P., O. T., M. W., M. G., R. M., V. H.;
Supervision and funding acquisition:    M. G., R. M., V. H.;
All authors have read and agreed to the published version of the article.

\printbibliography

\printacronyms

\section*{Appendix}
\label{sec:appendix}
\setcounter{figure}{0}
\setcounter{table}{0}
\setcounter{equation}{0}
\renewcommand{\thefigure}{A\arabic{table}}
\renewcommand{\thetable}{A\arabic{table}}
\renewcommand{\theequation}{A\arabic{table}}
\begin{table}[H]
    \caption{
        The configuration space $\mathbf{\Lambda}_\text{q}$ for the \acs{cINN}, which transforms a point forecasting model into a probabilistic one, in the respective naming convention~\cite{Phipps2023a}.
    }
    \label{tab:autopq_cinn-configurations}
    \centering
    \begin{adjustbox}{max width=\linewidth}
        \begin{tabular}{lll}
\toprule
\textbf{Hyperparameter}           & \textbf{Value range}                                  & \textbf{Default value}    \\
\midrule[\heavyrulewidth]\addlinespace
\texttt{sampling\_std}            & $[0.01, 3.0]$                                         & $0.1$           \\\addlinespace
\bottomrule
\end{tabular}%
    \end{adjustbox}
\end{table}

\begin{table}[H]
    \caption{
        The configuration of the \texttt{Propulate} \acs{EA} for \acs{HPO}, in the respective naming convention~\cite{Taubert2023}.
    }
    \label{tab:autopq_propulate-configuration}
    \centering
    \begin{adjustbox}{max width=\linewidth}
        \begin{tabular}{lll}
\toprule
\textbf{Module}                     & \textbf{Hyperparameter}             & \textbf{Value}  \\
\midrule[\heavyrulewidth]\addlinespace
\texttt{Islands}                    & \texttt{generations}                & -1              \\
                                    & \texttt{num\_isles}                 & 2               \\
                                    & \texttt{migration\_probability}     & 0.7             \\
                                    & \texttt{pollination}                & \texttt{True}   \\\addlinespace
\texttt{Compose}                    & \texttt{pop\_size}                  & 8               \\
                                    & \texttt{mate\_prob}                 & 0.7             \\
                                    & \texttt{mut\_prob}                  & 0.4             \\
                                    & \texttt{random\_prob}               & 0.2             \\
                                    & \texttt{sigma\_factor}              & 0.05            \\\addlinespace
\bottomrule
\end{tabular}%

    \end{adjustbox}
\end{table}

\begin{table}[H]
    \caption{
        The \acs{CRPS} evaluated on the hold\-/out test sub\-/sets before and after post\-/processing.
        In post\-/processing, negative values in the quantile forecasts are set to zero to maintain given limits (mobility indicator, \acs{PV} generation, and \acs{WP} generation are greater than or equal to zero).
        For the other three datasets (Load-BW, Load-\acs{GCP}, and Price), either no negative values exist in the quantile forecasts, respectively, negative values are valid.
    }
    \label{tab:autopq_post-processing}
    \begin{adjustbox}{max width=\linewidth}
\begin{tabular}{r|cc|l}
\cmidrule[\heavyrulewidth]{2-3}
& \textbf{AutoPQ-default} & \textbf{AutoPQ-advanced} &  \\
\cmidrule[\heavyrulewidth]{2-3}
Mobility
    & 0.2641 & 0.2585 & before post- \\[-2pt]
    & $\scriptscriptstyle\pm$\scriptsize0.0138 & $\scriptscriptstyle\pm$\scriptsize0.0051 & processing \\
    & 0.2627 & 0.2581 & after post- \\[-2pt]
    & $\scriptscriptstyle\pm$\scriptsize0.0037 & $\scriptscriptstyle\pm$\scriptsize0.0047 & processing \\
\cmidrule{2-3}
\acs{PV}
    & 0.1072 & 0.1018 & before post- \\[-2pt]
    & $\scriptscriptstyle\pm$\scriptsize0.0003 & $\scriptscriptstyle\pm$\scriptsize0.001 & processing \\
    & 0.1060 & 0.1013 & after post- \\[-2pt]
    & $\scriptscriptstyle\pm$\scriptsize0.0009 & $\scriptscriptstyle\pm$\scriptsize0.0007 & processing \\
\cmidrule{2-3}
\acs{WP}
    & 0.3798 & 0.3625 & before post- \\[-2pt]
    & $\scriptscriptstyle\pm$\scriptsize0.0018 & $\scriptscriptstyle\pm$\scriptsize0.0017 & processing \\
    & 0.3775 & 0.3621 & after post- \\[-2pt]
    & $\scriptscriptstyle\pm$\scriptsize0.0010 & $\scriptscriptstyle\pm$\scriptsize0.0012 & processing \\
\cmidrule[\heavyrulewidth]{2-3}
\end{tabular}%
    \end{adjustbox}
\end{table}

\begin{table*}
    \caption{
        The configuration spaces $\mathbf{\Lambda}_\text{p}$ for the \acs{SM}\-/based point forecasting methods in sktime naming convention~\cite{Löning2019} with the seasonal period $s=24$.
    }
    \label{tab:autopq_sm-configurations}
    \centering
    \begin{adjustbox}{max width=\linewidth}
        \begin{tabular}{llll}
\toprule
\textbf{Forecasting method} & \textbf{Hyperparameter}                   & \textbf{Value range}                  & \textbf{Default value}                        \\
\midrule[\heavyrulewidth]\addlinespace
\texttt{SARIMAX} \cite{Hyndman2021}             & \texttt{p}                                & $[0, 5]$                              & $1$                                           \\[-1pt]
                                                & \texttt{d}                                & $[0, 2]$                              & $0$                                           \\[-1pt]
                                                & \texttt{q}                                & $[0, 5]$                              & $0$                                           \\[-1pt]
                                                & \texttt{P}                                & $[0, 2]$                              & $0$                                           \\[-1pt]
                                                & \texttt{D}                                & $[0, 1]$                              & $0$                                           \\[-1pt]
                                                & \texttt{Q}                                & $[0, 2]$                              & $0$                                           \\\addlinespace
\texttt{ExponentialSmoothing} \cite{Hyndman2021}& \texttt{error}                            & $\{\texttt{add, mul}\}$               & $\texttt{add}$                                \\[-1pt]
                                                & \texttt{trend}                            & $\{\texttt{add, mul, None}\}$         & $\texttt{None}$                               \\[-1pt]
                                                & \texttt{seasonal}                         & $\{\texttt{add, mul, None}\}$         & $\texttt{None}$                               \\[-1pt]
                                                & \texttt{damped\_trend}                    & $\{\texttt{True, False}\}$            & $\texttt{False}$                              \\\addlinespace
\texttt{TBATS} \cite{DeLivera2011}              & \texttt{use\_trigonometric\_seasonality}  & $\{\texttt{True, False}\}$            & $\texttt{True}$                               \\[-1pt]
                                                & \texttt{use\_box\_cox}                    & $\{\texttt{True, False}\}$            & $\texttt{False}$                              \\[-1pt]
                                                & \texttt{use\_arma\_errors}                & $\{\texttt{True, False}\}$            & $\texttt{False}$                              \\[-1pt]
                                                & \texttt{use\_trend}                       & $\{\texttt{True, False}\}$            & $\texttt{False}$                              \\[-1pt]
                                                & \texttt{use\_damped\_trend}               & $\{\texttt{True, False}\}$            & $\texttt{False}$                              \\\addlinespace
\bottomrule
\end{tabular}%
    \end{adjustbox}
\end{table*}

\begin{table*}
    \caption{
        The configuration spaces $\mathbf{\Lambda}_\text{p}$ for the \acs{ML}\-/based point forecasting methods in the respective naming conventions.
    }
    \label{tab:autopq_ml-configurations}
    \centering
    \begin{adjustbox}{max width=\linewidth}
        \begin{tabular}{llll}
\toprule
\textbf{Forecasting method} & \textbf{Hyperparameter}       & \textbf{Value range}                                  & \textbf{Default value}    \\
\midrule[\heavyrulewidth]\addlinespace
\texttt{MLPRegressor} \cite{Pedregosa2011}      & \texttt{activation}           & $\{\texttt{logistic, tanh, relu}\}$                   & $\texttt{relu}$                   \\[-1pt]
                                                & \texttt{batch\_size}          & $\{32, 64, 128\}$                                     & $\min(200, \texttt{n\_samples})$                  \\[-1pt]
                                                & \texttt{hidden\_layer\_sizes} & $\{([10, 100]),$                                      & $(100)$                           \\[-1pt]
                                                &                               & $\phantom{\{}([10, 100], [10, 100]),$                 &                                   \\[-1pt]
                                                &                               & $\phantom{\{}([10, 100], [10, 100], [10, 100])\}$     &                                   \\\addlinespace
\texttt{MSVR} \cite{Bao2014}                    & \texttt{C}                    & $[0.01, 100]$                                         & $1.0$                             \\[-1pt]
                                                & \texttt{epsilon}              & $[0.001, 1]$                                          & $0.1$                             \\[-1pt]
                                                & \texttt{kernel}               & $\{\texttt{rbf, laplacian, sigmoid}\}$                & $\texttt{rbf}$                    \\\addlinespace
\texttt{XGBRegressor} \cite{Chen2016}           & \texttt{learning\_rate}       & $[0.01, 1]$                                           & $0.3$                             \\[-1pt]
                                                & \texttt{max\_depth}           & $[1, 18]$                                             & $6$                               \\[-1pt]
                                                & \texttt{n\_estimators}        & $[10, 300]$                                           & $100$                             \\[-1pt]
                                                & \texttt{sub\_sample}          & $[0.5, 1.0]$                                          & $1.0$                             \\\addlinespace
\bottomrule
\end{tabular}%
    \end{adjustbox}
\end{table*}

\begin{table*}
    \caption{
        The configuration spaces $\mathbf{\Lambda}_\text{p}$ for the \acs{DL}\-/based (point) forecasting methods in PyTorch Forecasting naming convention~\cite{Beitner2020}.
    }
    \label{tab:autopq_dl-configurations}
    \centering
    \begin{adjustbox}{max width=\linewidth}
        \begin{tabular}{llll}
\toprule
\textbf{Forecasting method} & \textbf{Hyperparameter}           & \textbf{Value range}                                  & \textbf{Default value}    \\
\midrule[\heavyrulewidth]\addlinespace
\texttt{DeepAR} \cite{Salinas2020}              & \texttt{batch\_size}              & $\{32, 64, 128\}$                                     & $64$                      \\[-1pt]
                                                & \texttt{cell\_type}               & $\{\texttt{LSTM, GRU}\}$                              & $\texttt{LSTM}$           \\[-1pt]
                                                & \texttt{dropout}                  & $[0.0, 0.2]$                                          & $0.1$                     \\[-1pt]
                                                & \texttt{hidden\_size}             & $[10, 100]$                                           & $10$                      \\[-1pt]
                                                & \texttt{rnn\_layers}              & $[1,3]$                                               & $2$                       \\\addlinespace
\texttt{NHiTS} \cite{Challu2023}                & \texttt{batch\_size}              & $\{32, 64, 128\}$                                     & $64$                      \\[-1pt]
                                                & \texttt{dropout}                  & $[0.0, 0.2]$                                          & $0.0$                     \\[-1pt]
                                                & \texttt{hidden\_size}             & $[8, 1024]$                                           & $512$                     \\[-1pt]
                                                & \texttt{n\_blocks}                & $\{[1],[1,1],[1,1,1]\}$                               & $[1, 1, 1]$               \\[-1pt]
                                                & \texttt{n\_layers}                & $[1,3]$                                               & $2$                       \\[-1pt]
                                                & \texttt{shared\_weights}          & $\{\texttt{True, False}\}$                            & $\texttt{True}$           \\\addlinespace
\texttt{TemporalFusion} -                       & \texttt{batch\_size}              & $\{32, 64, 128\}$                                     & $64$                      \\[-1pt]
\texttt{\quad Transformer} \cite{Lim2021b}      & \texttt{dropout}                  & $[0.0, 0.2]$                                          & $0.1$                     \\[-1pt]
                                                & \texttt{hidden\_continuous\_size} & $[8, 64]$                                             & $8$                       \\[-1pt]
                                                & \texttt{hidden\_size}             & $[16, 256]$                                           & $16$                      \\[-1pt]
                                                & \texttt{lstm\_layers}             & $[1,3]$                                               & $1$                       \\\addlinespace
\bottomrule
\end{tabular}%
    \end{adjustbox}
    \scriptsize\\[0.5mm]
    \acs{DeepAR} may also provide probabilistic forecasts.
\end{table*}

\begin{table*}
    \caption{
        Overview of the training, validation, and test sub\-/sets of the data with hourly resolution used for evaluation, as well as the selected features based on~\cite{Phipps2024}.
        The names of the target variables and the features refer to the column names in the datasets.
    }
    \label{tab:autopq_data}
    \centering
    \begin{adjustbox}{max width=\linewidth}
        \begin{tabular}{lllccc}
\toprule
\textbf{Data set}   & \textbf{Target variable}  & \textbf{Features}     & \textbf{Training sub-set} & \textbf{Validation sub-set}   & \textbf{Test sub-set} \\
\midrule[\heavyrulewidth]
Load-BW
& \texttt{load\_power\_statistics}
& \texttt{Seasonal features} \eqref{eq:concept_trig-12}, \eqref{eq:concept_trig-24}, \eqref{eq:concept_workday-weekend-holiday}
& [0,4904]                  & [4905,7007]                    & [7008,8760]           \\
&
& \texttt{Lag features} \eqref{eq:concept_lag}
& (\SI{204.3}{\day})        & (\SI{87.6}{\day})              & (\SI{73.0}{\day})     \\\addlinespace
Load-\acs{GCP}
& \texttt{MT\_158}
& \texttt{Seasonal features} \eqref{eq:concept_trig-12}, \eqref{eq:concept_trig-24}, \eqref{eq:concept_workday-weekend-holiday}
& [0,14716]                 & [14717,21023]                 & [21024,26280]         \\
&                           & \texttt{Lag features} \eqref{eq:concept_lag}
& (\SI{613.2}{\day})        & (\SI{262.8}{\day})                              & (\SI{219.0}{\day})                      \\\addlinespace
Mobility
& \texttt{cnt}
& \texttt{Seasonal features} \eqref{eq:concept_trig-12}, \eqref{eq:concept_trig-24}, \eqref{eq:concept_workday-weekend-holiday}
& [0,9824]                  & [9825,14034]                  & [14035,17544]         \\
&                           & \texttt{Lag features} \eqref{eq:concept_lag}
& (\SI{409.3}{\day})                          & (\SI{175.4}{\day})                              & (\SI{146.2}{\day})                      \\
&                           & \texttt{temp}
&                           &                               &                       \\
&                           & \texttt{hum}
&                           &                               &                       \\
&                           & \texttt{windspeed}
&                           &                               &                       \\
&                           & \texttt{weathersit}
&                           &                               &                       \\\addlinespace
Price
& \texttt{Zonal Price}
& \texttt{Seasonal features} \eqref{eq:concept_trig-12}, \eqref{eq:concept_trig-24}, \eqref{eq:concept_workday-weekend-holiday}
& [0,14541]                 & [14542,20773]                 & [20774,25968]         \\
&                           & \texttt{Lag features} \eqref{eq:concept_lag}
& (\SI{605.9}{\day})                          & (\SI{259.6}{\day})                              & (\SI{216.4}{\day})                      \\
&                           & \texttt{Forecast Total Load}
&                           &                               &                       \\
&                           & \texttt{Forecast Zonal Load}
&                           &                               &                       \\\addlinespace
\acs{PV}
& \texttt{POWER}                     
& \texttt{Seasonal features} \eqref{eq:concept_trig-12}, \eqref{eq:concept_trig-24}, \eqref{eq:concept_workday-weekend-holiday}
& [0,11033]                 & [11034,15762]                 & [15763,19704]         \\
&                           & \texttt{Lag features} \eqref{eq:concept_lag}
& (\SI{459.7}{\day})                          & (\SI{197.0}{\day})                              & (\SI{164.2}{\day})                      \\
&                           & \texttt{SSRD}
&                           &                               &                       \\
&                           & \texttt{TCC}
&                           &                               &                       \\\addlinespace
\acs{WP}
& \texttt{TARGETVAR}
& \texttt{Seasonal features} \eqref{eq:concept_trig-12}, \eqref{eq:concept_trig-24}, \eqref{eq:concept_workday-weekend-holiday}
& [0,9400]                  & [9401,13430]                  & [13431,16789]         \\
&                           & \texttt{Lag features} \eqref{eq:concept_lag}
& (\SI{391.7}{\day})                          & (\SI{167.9}{\day})                              & (\SI{139.9}{\day})                      \\
&                           & \texttt{U100}
&                           &                               &                       \\
&                           & \texttt{V100}
&                           &                               &                       \\
&                           & \texttt{Speed100}
&                           &                               &                       \\
\bottomrule
\end{tabular}
    \end{adjustbox}
    \scriptsize\\[0.5mm]
    \texttt{cnt}: count; \texttt{temp}: air temperature; \texttt{hum}: humidity; \texttt{weathersit}: weather situation; \texttt{SSRD}: Surface Short-wave (solar) Radiation Downwards; \texttt{TCC}: Total Cloud Cover; \texttt{U100}, \texttt{V100}, \texttt{Speed100}: wind speeds at 100 meters above ground, with the eastward component (U), the northward component (V), and the total wind speed (Speed).
\end{table*}


\begin{table*}
    \caption{
        The validation \acs{CRPS} compared for three trial generators in the inner loop of \autoref{alg:autopq_combined-smart} (random search, \acs{BO}-\acs{TPE}, \acs{BO}-\acs{TPE}-\acs{PK}).
        Across the six datasets and the nine forecasting methods, only minor differences occur.
    }
    \label{tab:autopq_n-iter_score}
    \begin{adjustbox}{max width=\linewidth}
\begin{tabular}{r|ccc|ccc|ccc|l}
\cmidrule[\heavyrulewidth]{2-10}
& \textbf{\acs{ETS}} & \textbf{\acs{sARIMAX}} & \textbf{\acs{TBATS}} & \textbf{\acs{MLP}} & \textbf{\acs{SVR}} & \textbf{\acs{XGB}} & \textbf{\acs{DeepAR}} & \textbf{\acs{NHiTS}} & \textbf{\acs{TFT}} &  \\
\cmidrule[\heavyrulewidth]{2-10}
Load-BW & 0.357 & 0.266 & 0.259 & 0.099 & 0.100 & 0.079 & 0.134 & 0.090 & 0.107 & random \\[-2pt]
      & $\scriptscriptstyle\pm$\scriptsize0.018 & $\scriptscriptstyle\pm$\scriptsize0.003 & $\scriptscriptstyle\pm$\scriptsize0.004 & $\scriptscriptstyle\pm$\scriptsize0.008 & $\scriptscriptstyle\pm$\scriptsize0.001 & $\scriptscriptstyle\pm$\scriptsize0.001 & $\scriptscriptstyle\pm$\scriptsize0.01 & $\scriptscriptstyle\pm$\scriptsize0.004 & $\scriptscriptstyle\pm$\scriptsize0.006 & search \\
      & 0.357 & 0.266 & 0.259 & 0.103 & 0.100 & 0.079 & 0.133 & 0.090 & 0.107 & \acs{BO} \\[-2pt]
      & $\scriptscriptstyle\pm$\scriptsize0.018 & $\scriptscriptstyle\pm$\scriptsize0.003 & $\scriptscriptstyle\pm$\scriptsize0.004 & $\scriptscriptstyle\pm$\scriptsize0.011 & $\scriptscriptstyle\pm$\scriptsize0.001 & $\scriptscriptstyle\pm$\scriptsize0.001 & $\scriptscriptstyle\pm$\scriptsize0.01 & $\scriptscriptstyle\pm$\scriptsize0.004 & $\scriptscriptstyle\pm$\scriptsize0.006 & \acs{TPE} \\
      & 0.357 & 0.277 & 0.259 & 0.100 & 0.101 & 0.079 & 0.133 & 0.089 & 0.107 & \acs{BO} \\[-2pt]
      & $\scriptscriptstyle\pm$\scriptsize0.018 & $\scriptscriptstyle\pm$\scriptsize0.014 & $\scriptscriptstyle\pm$\scriptsize0.004 & $\scriptscriptstyle\pm$\scriptsize0.008 & $\scriptscriptstyle\pm$\scriptsize0.001 & $\scriptscriptstyle\pm$\scriptsize0.001 & $\scriptscriptstyle\pm$\scriptsize0.01 & $\scriptscriptstyle\pm$\scriptsize0.002 & $\scriptscriptstyle\pm$\scriptsize0.006 & \acs{TPE} \acs{PK} \\
\cmidrule{2-10}
Load-\acs{GCP} & 0.576 & 0.491 & 0.405 & 0.219 & 0.215 & 0.197 & 0.247 & 0.230 & 0.212 & random \\[-2pt]
      & $\scriptscriptstyle\pm$\scriptsize0.024 & $\scriptscriptstyle\pm$\scriptsize0.005 & $\scriptscriptstyle\pm$\scriptsize0.008 & $\scriptscriptstyle\pm$\scriptsize0.002 & $\scriptscriptstyle\pm$\scriptsize0.002 & $\scriptscriptstyle\pm$\scriptsize0.001 & $\scriptscriptstyle\pm$\scriptsize0.007 & $\scriptscriptstyle\pm$\scriptsize0.004 & $\scriptscriptstyle\pm$\scriptsize0.002 & search \\
      & 0.576 & 0.491 & 0.405 & 0.219 & 0.215 & 0.197 & 0.247 & 0.230 & 0.212 & \acs{BO} \\[-2pt]
      & $\scriptscriptstyle\pm$\scriptsize0.024 & $\scriptscriptstyle\pm$\scriptsize0.005 & $\scriptscriptstyle\pm$\scriptsize0.008 & $\scriptscriptstyle\pm$\scriptsize0.002 & $\scriptscriptstyle\pm$\scriptsize0.002 & $\scriptscriptstyle\pm$\scriptsize0.001 & $\scriptscriptstyle\pm$\scriptsize0.007 & $\scriptscriptstyle\pm$\scriptsize0.004 & $\scriptscriptstyle\pm$\scriptsize0.002 & \acs{TPE} \\
      & 0.576 & 0.491 & 0.405 & 0.219 & 0.215 & 0.197 & 0.247 & 0.230 & 0.212 & \acs{BO} \\[-2pt]
      & $\scriptscriptstyle\pm$\scriptsize0.024 & $\scriptscriptstyle\pm$\scriptsize0.005 & $\scriptscriptstyle\pm$\scriptsize0.008 & $\scriptscriptstyle\pm$\scriptsize0.002 & $\scriptscriptstyle\pm$\scriptsize0.002 & $\scriptscriptstyle\pm$\scriptsize0.001 & $\scriptscriptstyle\pm$\scriptsize0.007 & $\scriptscriptstyle\pm$\scriptsize0.004 & $\scriptscriptstyle\pm$\scriptsize0.002 & \acs{TPE} \acs{PK} \\
\cmidrule{2-10}
Mobility & 0.630 & 0.611 & 0.446 & 0.377 & 0.416 & 0.397 & 0.280 & 0.295 & 0.261 & random \\[-2pt]
      & $\scriptscriptstyle\pm$\scriptsize0.007 & $\scriptscriptstyle\pm$\scriptsize0.01 & $\scriptscriptstyle\pm$\scriptsize0.001 & $\scriptscriptstyle\pm$\scriptsize0.014 & $\scriptscriptstyle\pm$\scriptsize0.004 & $\scriptscriptstyle\pm$\scriptsize0.005 & $\scriptscriptstyle\pm$\scriptsize0.01 & $\scriptscriptstyle\pm$\scriptsize0.05 & $\scriptscriptstyle\pm$\scriptsize0.008 & search \\
      & 0.630 & 0.611 & 0.446 & 0.376 & 0.416 & 0.397 & 0.279 & 0.294 & 0.261 & \acs{BO} \\[-2pt]
      & $\scriptscriptstyle\pm$\scriptsize0.008 & $\scriptscriptstyle\pm$\scriptsize0.01 & $\scriptscriptstyle\pm$\scriptsize0.002 & $\scriptscriptstyle\pm$\scriptsize0.006 & $\scriptscriptstyle\pm$\scriptsize0.004 & $\scriptscriptstyle\pm$\scriptsize0.005 & $\scriptscriptstyle\pm$\scriptsize0.011 & $\scriptscriptstyle\pm$\scriptsize0.048 & $\scriptscriptstyle\pm$\scriptsize0.008 & \acs{TPE} \\
      & 0.630 & 0.611 & 0.445 & 0.374 & 0.416 & 0.397 & 0.280 & 0.288 & 0.261 & \acs{BO} \\[-2pt]
      & $\scriptscriptstyle\pm$\scriptsize0.007 & $\scriptscriptstyle\pm$\scriptsize0.01 & $\scriptscriptstyle\pm$\scriptsize0.002 & $\scriptscriptstyle\pm$\scriptsize0.006 & $\scriptscriptstyle\pm$\scriptsize0.005 & $\scriptscriptstyle\pm$\scriptsize0.005 & $\scriptscriptstyle\pm$\scriptsize0.01 & $\scriptscriptstyle\pm$\scriptsize0.036 & $\scriptscriptstyle\pm$\scriptsize0.007 & \acs{TPE} \acs{PK} \\
\cmidrule{2-10}
Price & 0.551 & 0.252 & 0.285 & 0.194 & 0.205 & 0.206 & 0.189 & 0.164 & 0.194 & random \\[-2pt]
      & $\scriptscriptstyle\pm$\scriptsize0.012 & $\scriptscriptstyle\pm$\scriptsize0.006 & $\scriptscriptstyle\pm$\scriptsize0.003 & $\scriptscriptstyle\pm$\scriptsize0.002 & $\scriptscriptstyle\pm$\scriptsize0.001 & $\scriptscriptstyle\pm$\scriptsize0.003 & $\scriptscriptstyle\pm$\scriptsize0.01 & $\scriptscriptstyle\pm$\scriptsize0.01 & $\scriptscriptstyle\pm$\scriptsize0.008 & search \\
      & 0.550 & 0.252 & 0.285 & 0.196 & 0.205 & 0.206 & 0.190 & 0.162 & 0.194 & \acs{BO} \\[-2pt]
      & $\scriptscriptstyle\pm$\scriptsize0.011 & $\scriptscriptstyle\pm$\scriptsize0.006 & $\scriptscriptstyle\pm$\scriptsize0.003 & $\scriptscriptstyle\pm$\scriptsize0.004 & $\scriptscriptstyle\pm$\scriptsize0.001 & $\scriptscriptstyle\pm$\scriptsize0.003 & $\scriptscriptstyle\pm$\scriptsize0.01 & $\scriptscriptstyle\pm$\scriptsize0.007 & $\scriptscriptstyle\pm$\scriptsize0.008 & \acs{TPE} \\
      & 0.550 & 0.252 & 0.285 & 0.194 & 0.205 & 0.206 & 0.191 & 0.162 & 0.194 & \acs{BO} \\[-2pt]
      & $\scriptscriptstyle\pm$\scriptsize0.012 & $\scriptscriptstyle\pm$\scriptsize0.006 & $\scriptscriptstyle\pm$\scriptsize0.003 & $\scriptscriptstyle\pm$\scriptsize0.005 & $\scriptscriptstyle\pm$\scriptsize0.001 & $\scriptscriptstyle\pm$\scriptsize0.003 & $\scriptscriptstyle\pm$\scriptsize0.003 & $\scriptscriptstyle\pm$\scriptsize0.007 & $\scriptscriptstyle\pm$\scriptsize0.008 & \acs{TPE} \acs{PK} \\
\cmidrule{2-10}
\acs{PV}    & 0.303 & 0.696 & 0.165 & 0.127 & 0.114 & 0.107 & 0.158 & 0.161 & 0.131 & random \\[-2pt]
      & $\scriptscriptstyle\pm$\scriptsize0.018 & $\scriptscriptstyle\pm$\scriptsize0.196 & $\scriptscriptstyle\pm$\scriptsize0.001 & $\scriptscriptstyle\pm$\scriptsize0.002 & $\scriptscriptstyle\pm$\scriptsize0.001 & $\scriptscriptstyle\pm$\scriptsize0.001 & $\scriptscriptstyle\pm$\scriptsize0.004 & $\scriptscriptstyle\pm$\scriptsize0.025 & $\scriptscriptstyle\pm$\scriptsize0.001 & search \\
      & 0.303 & 0.695 & 0.165 & 0.127 & 0.114 & 0.107 & 0.158 & 0.168 & 0.131 & \acs{BO} \\[-2pt]
      & $\scriptscriptstyle\pm$\scriptsize0.018 & $\scriptscriptstyle\pm$\scriptsize0.196 & $\scriptscriptstyle\pm$\scriptsize0.001 & $\scriptscriptstyle\pm$\scriptsize0.002 & $\scriptscriptstyle\pm$\scriptsize0 & $\scriptscriptstyle\pm$\scriptsize0.001 & $\scriptscriptstyle\pm$\scriptsize0.004 & $\scriptscriptstyle\pm$\scriptsize0.043 & $\scriptscriptstyle\pm$\scriptsize0.001 & \acs{TPE} \\
      & 0.303 & 0.695 & 0.165 & 0.127 & 0.114 & 0.107 & 0.158 & 0.159 & 0.131 & \acs{BO} \\[-2pt]
      & $\scriptscriptstyle\pm$\scriptsize0.017 & $\scriptscriptstyle\pm$\scriptsize0.196 & $\scriptscriptstyle\pm$\scriptsize0.001 & $\scriptscriptstyle\pm$\scriptsize0.002 & $\scriptscriptstyle\pm$\scriptsize0 & $\scriptscriptstyle\pm$\scriptsize0.001 & $\scriptscriptstyle\pm$\scriptsize0.005 & $\scriptscriptstyle\pm$\scriptsize0.023 & $\scriptscriptstyle\pm$\scriptsize0.001 & \acs{TPE} \acs{PK} \\
\cmidrule{2-10}
\acs{WP}    & 0.549 & 0.497 & 0.547 & 0.320 & 0.303 & 0.292 & 0.412 & 0.389 & 0.407 & random \\[-2pt]
      & $\scriptscriptstyle\pm$\scriptsize0.004 & $\scriptscriptstyle\pm$\scriptsize0.008 & $\scriptscriptstyle\pm$\scriptsize0.011 & $\scriptscriptstyle\pm$\scriptsize0.009 & $\scriptscriptstyle\pm$\scriptsize0.003 & $\scriptscriptstyle\pm$\scriptsize0.001 & $\scriptscriptstyle\pm$\scriptsize0.003 & $\scriptscriptstyle\pm$\scriptsize0.011 & $\scriptscriptstyle\pm$\scriptsize0.017 & search \\
      & 0.549 & 0.496 & 0.547 & 0.321 & 0.303 & 0.292 & 0.412 & 0.389 & 0.408 & \acs{BO} \\[-2pt]
      & $\scriptscriptstyle\pm$\scriptsize0.004 & $\scriptscriptstyle\pm$\scriptsize0.008 & $\scriptscriptstyle\pm$\scriptsize0.012 & $\scriptscriptstyle\pm$\scriptsize0.013 & $\scriptscriptstyle\pm$\scriptsize0.003 & $\scriptscriptstyle\pm$\scriptsize0.001 & $\scriptscriptstyle\pm$\scriptsize0.003 & $\scriptscriptstyle\pm$\scriptsize0.011 & $\scriptscriptstyle\pm$\scriptsize0.017 & \acs{TPE} \\
      & 0.549 & 0.502 & 0.545 & 0.319 & 0.303 & 0.292 & 0.412 & 0.390 & 0.407 & \acs{BO} \\[-2pt]
      & $\scriptscriptstyle\pm$\scriptsize0.003 & $\scriptscriptstyle\pm$\scriptsize0.019 & $\scriptscriptstyle\pm$\scriptsize0.013 & $\scriptscriptstyle\pm$\scriptsize0.005 & $\scriptscriptstyle\pm$\scriptsize0.003 & $\scriptscriptstyle\pm$\scriptsize0.001 & $\scriptscriptstyle\pm$\scriptsize0.004 & $\scriptscriptstyle\pm$\scriptsize0.01 & $\scriptscriptstyle\pm$\scriptsize0.017 & \acs{TPE} \acs{PK} \\
\cmidrule[\heavyrulewidth]{2-10}
\end{tabular}%
    \end{adjustbox}
\end{table*}

\end{document}